%% file: arxiv.tex
\documentclass[10pt,twocolumn,letterpaper]{article}

\usepackage{cvpr}
\usepackage{times}
\usepackage{epsfig}
\usepackage{graphicx}
\usepackage{amsmath}
\usepackage[usenames,dvipsnames,svgnames,table]{xcolor}
\usepackage{amssymb}
\usepackage{multirow}
\usepackage{subcaption}
\usepackage{verbatim}
\usepackage{epigraph}	

\usepackage[export]{adjustbox}

\usepackage[pagebackref=true,breaklinks=true,letterpaper=true,colorlinks,bookmarks=false]{hyperref}

\makeatletter
\usepackage{xspace}
\def\@onedot{\ifx\@let@token.\else.\null\fi\xspace}
\DeclareRobustCommand\onedot{\futurelet\@let@token\@onedot}

\newcommand{\figref}[1]{Fig\onedot~\ref{#1}}

\def\etc{\emph{etc}\onedot}

\makeatother

\cvprfinalcopy 

\def\cvprPaperID{2228} 

\ifcvprfinal\fi
\begin{document}

\title{TorontoCity: Seeing the World with a Million Eyes}

\author{\small Shenlong Wang, Min Bai, Gellert Mattyus, Hang Chu, Wenjie Luo, Bin Yang, Justin Liang, Joel Cheverie, Sanja Fidler,  Raquel Urtasun\\
\small Department of Computer Science, University of Toronto \\
{\tt\scriptsize \{slwang, mbai, mattyusg, chuhang1122, wenjie, byang, justinliang, joel, fidler, urtasun\}@cs.toronto.edu}
}
\newcommand{\high}[1]{{\color{blue}{#1}}}
\newcommand{\raquel}[1]{{\color{red}{#1}}}
\newcommand{\shenlong}[1]{{\color{black}{#1}}}
\newcommand{\joeljustin}[1]{{\color{black}{#1}}}

\newcommand{\bx}{{\mathbf{x}}}
\newcommand{\by}{{\mathbf{y}}}

\maketitle
\input{abstract}

\input{intro}

\input{related}

\input{dataset}
\input{method}

\input{benchmarks}
\input{results}
\input{conc}

{\small
\bibliographystyle{ieee}
\bibliography{egbib}
}

\end{document}

%% file: abstract.tex
 
\begin{abstract}

In this paper we introduce the 
TorontoCity benchmark, which covers 
the full greater Toronto area (GTA) with $712.5km^2$ of land, $8439km$ of road and around $400,000$ buildings. 
Our benchmark provides \joeljustin{different perspectives} 
of the world captured from airplanes, drones \joeljustin{and} 
cars driving around the city. 
Manually labeling such a large scale dataset is infeasible. 
Instead,  we propose to utilize different sources of high-precision maps  to create our ground truth.
Towards this goal, we develop algorithms that allow us to align all  data sources with the maps while requiring  minimal human supervision. 
We have designed a wide variety of tasks including building height estimation (reconstruction), road centerline and curb extraction, building instance segmentation, building contour extraction (reorganization), semantic labeling and scene type classification (recognition). 
Our pilot study shows that most of these tasks are still \joeljustin{difficult} for modern convolutional \joeljustin{neural} networks. 

\end{abstract}

%% file: intro.tex
\section{Introduction} 

\epigraph{"It is a narrow mind which cannot look at a subject from various points of view."}{\textit{George Eliot,  Middlemarch}}


In recent times, a great deal of effort has
been devoted to creating large scale benchmarks. These have been instrumental to the development of the field, and have enabled many significant break-throughs.
ImageNet \cite{imagenet} made it possible to train large-scale convolutional neural networks, initiating the deep learning  revolution in computer vision in 2012 with SuperVision (most commonly refer as AlexNet \cite{alexnet}).
Efforts 
such as PASCAL \cite{pascal} and  Microsoft COCO \cite{coco}  have pushed the performance of  segmentation and object detection approaches to 
previously inconceivable levels. 
Similarly, benchmarks 
such as KITTI \cite{kitti} and Cityscapes \cite{cityscape} have shown that visual perception is going to be an important component of advanced driver assistance systems (ADAS) 
and self-driving cars in the 
imminent 
future. 


 \begin{figure*}[t]
  \vspace{-0.5cm}
  \begin{minipage}{0.02\textwidth}
   \rotatebox{90}{Data}
   \rotatebox{90}{Maps\ \ \ \ \ \ \ \ \ \ \ \ \ \ \ \ \ \ \ \ \ \ \ \ \ \ \ \ \ \ \ \ \ \ \ }
   \rotatebox{90}{Tasks\ \ \ \ \ \ \ \ \ \ \ \ \ \ \ \ \ \ \ \ \ \ \ \ \ \ \ \ \ \ \ \ \ }
    \end{minipage}
 \begin{minipage}{0.98\textwidth}
 \begin{center}
   \begin{subfigure}[t]{\textwidth}
 \includegraphics[width=0.998\linewidth]{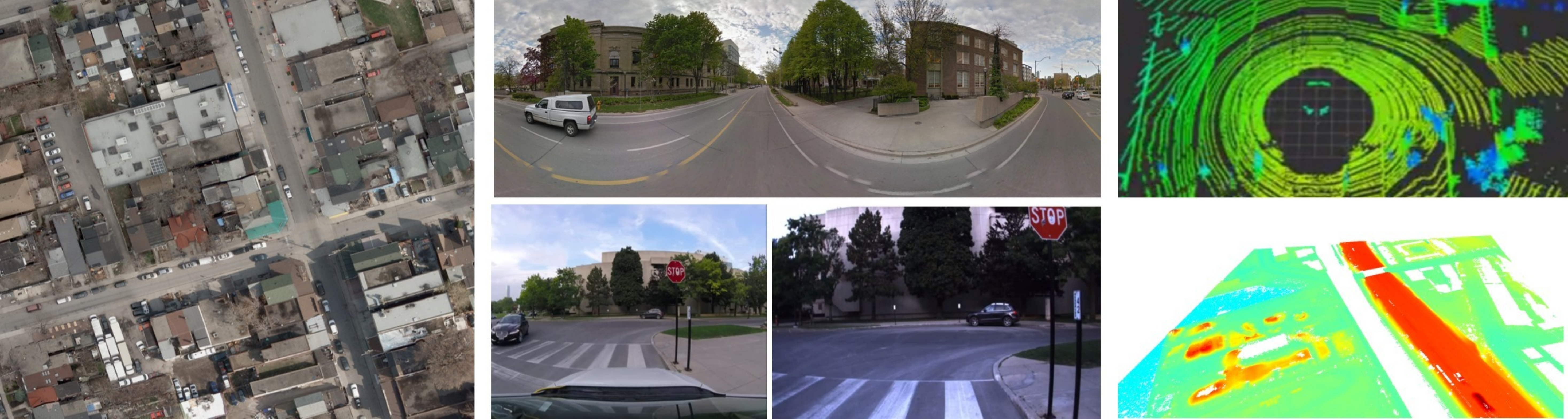}
 \label{fig:data}
   \end{subfigure}
   
   \vspace{-3mm}
   \noindent\makebox[\linewidth]{\rule{\linewidth}{1pt}}
   
  \begin{subfigure}[t]{\textwidth}
 \includegraphics[width=0.998\linewidth]{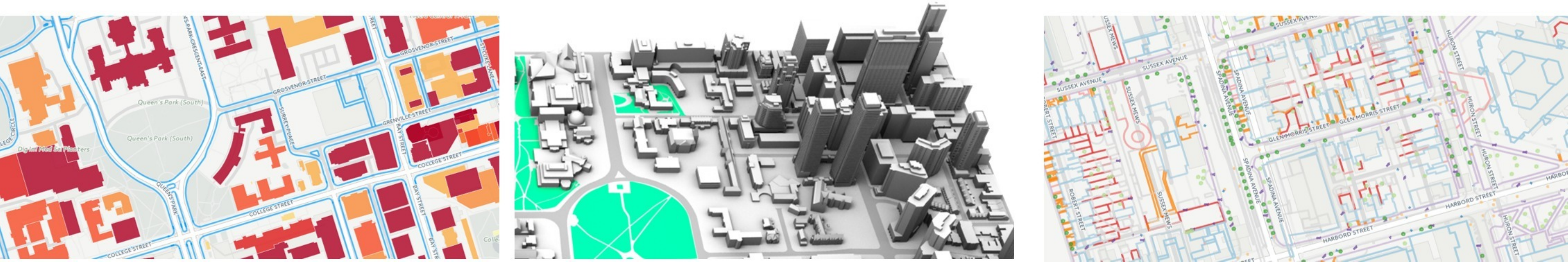}
 \label{fig:map}
\end{subfigure}

   \vspace{-3mm}
\noindent\makebox[\linewidth]{\rule{\linewidth}{1pt}}

   \vspace{1mm}

\begin{subfigure}[t]{\textwidth}
   \centering
 \adjincludegraphics[width=0.196\linewidth,trim={0 0 {.3\width} {.1\width}},clip]{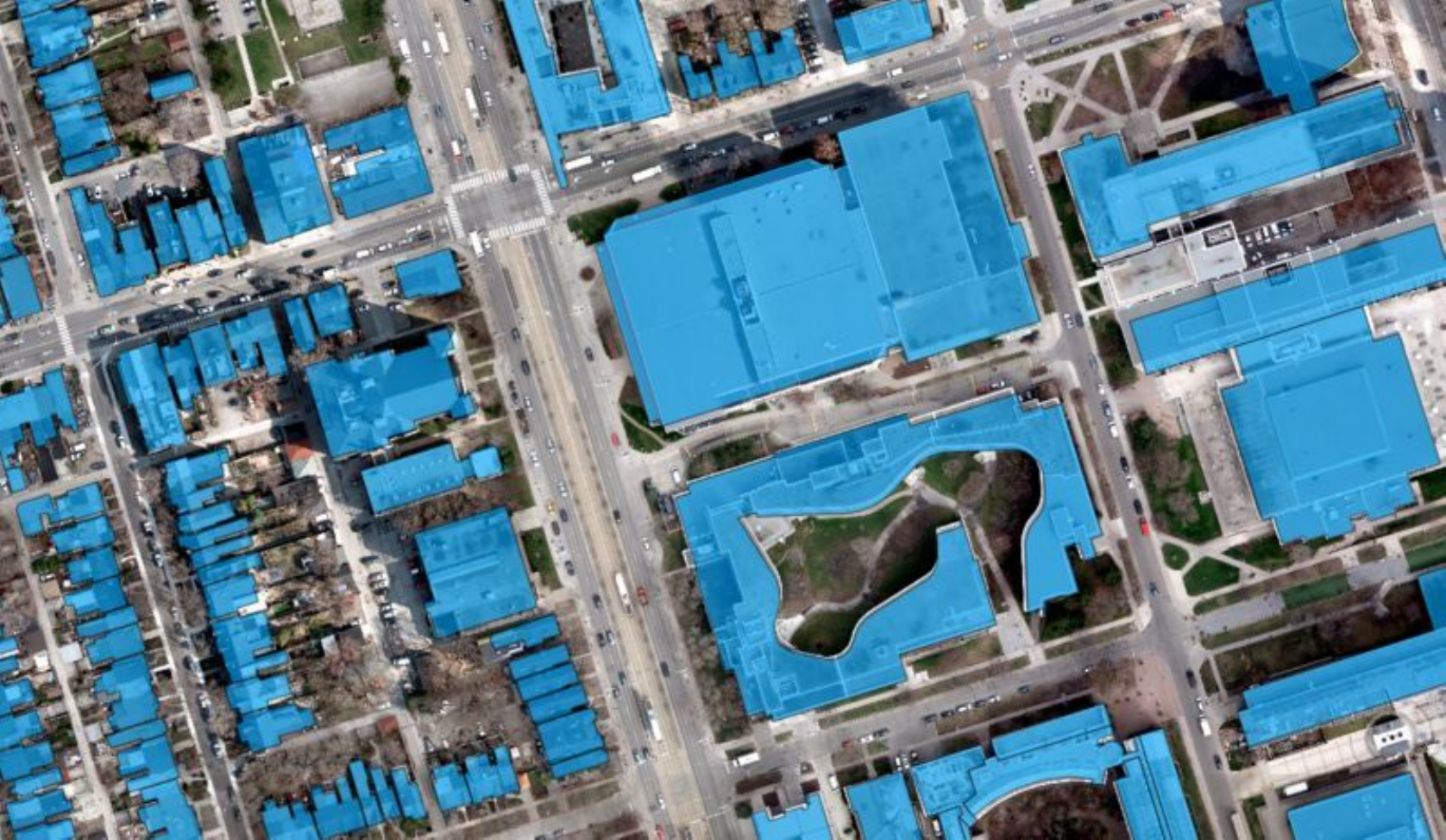}\ 
 \adjincludegraphics[width=0.196\linewidth,trim={0 0 {.3\width} {.1\width}},clip]{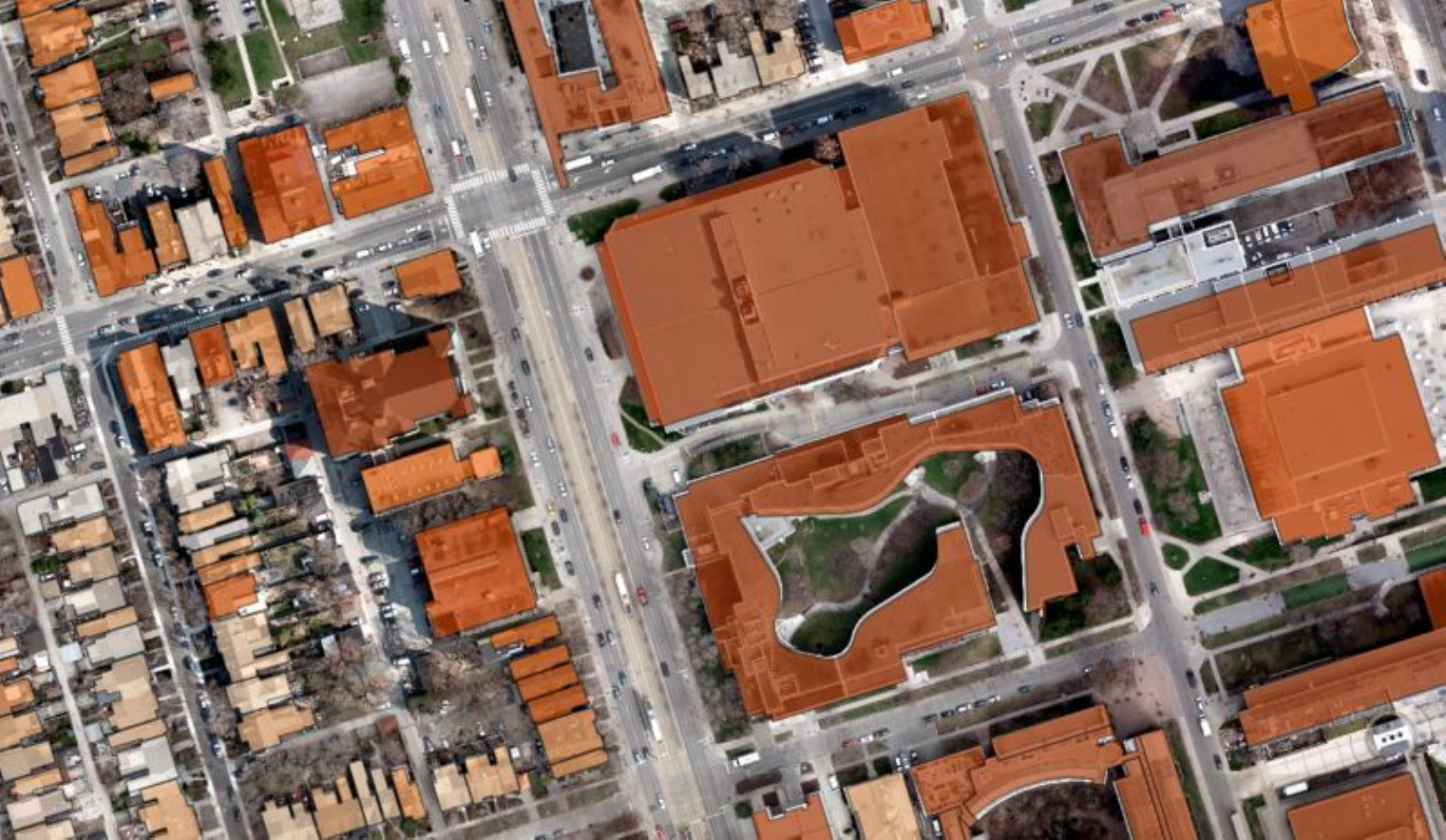}\ 
 \adjincludegraphics[width=0.196\linewidth,trim={0 0 {.3\width} {.1\width}},clip]{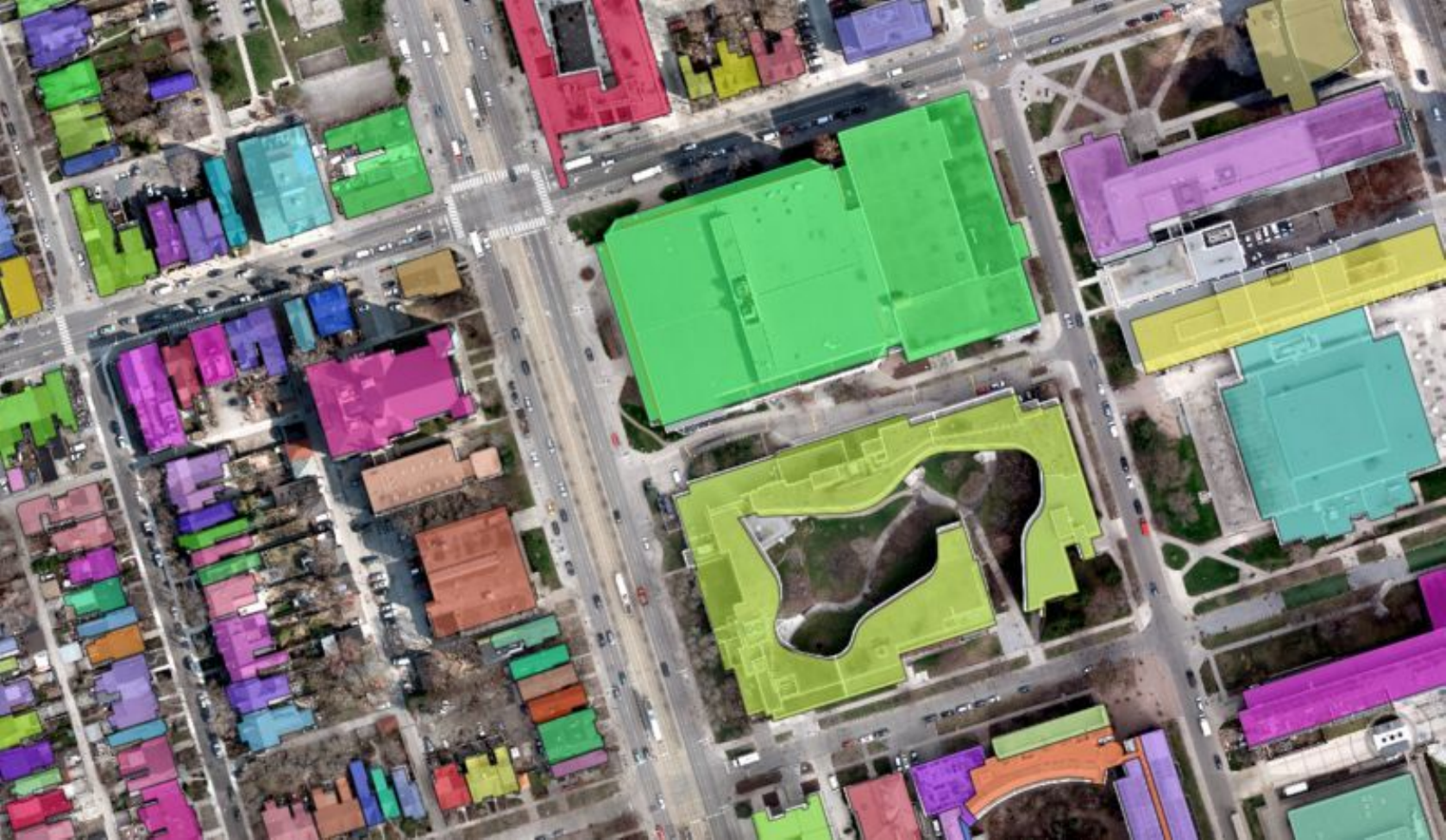}\ 
 \adjincludegraphics[width=0.196\linewidth,trim={0 0 {.3\width} {.1\width}},clip]{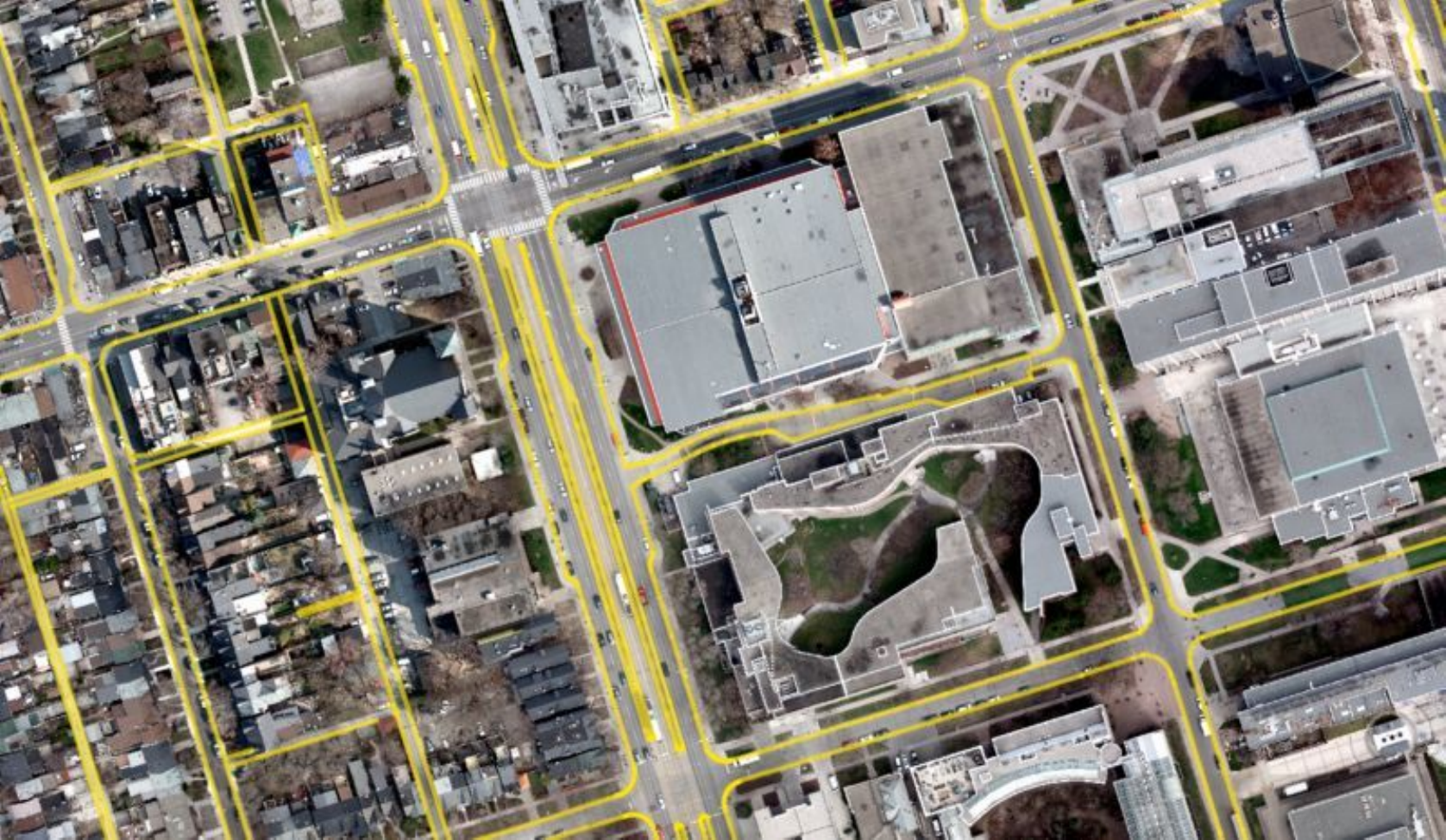}\ 
 \adjincludegraphics[width=0.196\linewidth,trim={0 0 {.3\width} {.1\width}},clip]{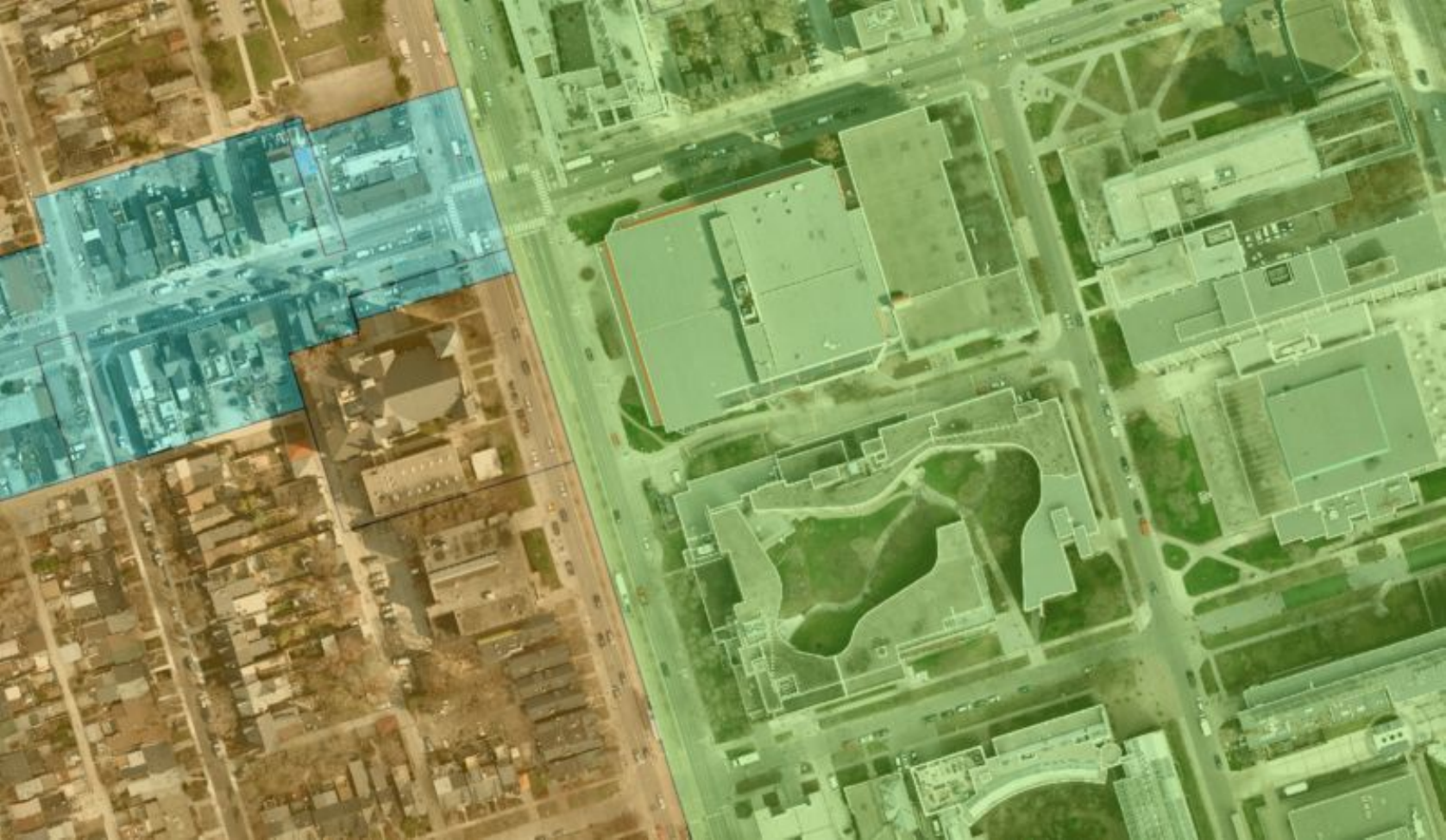}\ 
  \includegraphics[width=0.246\linewidth]{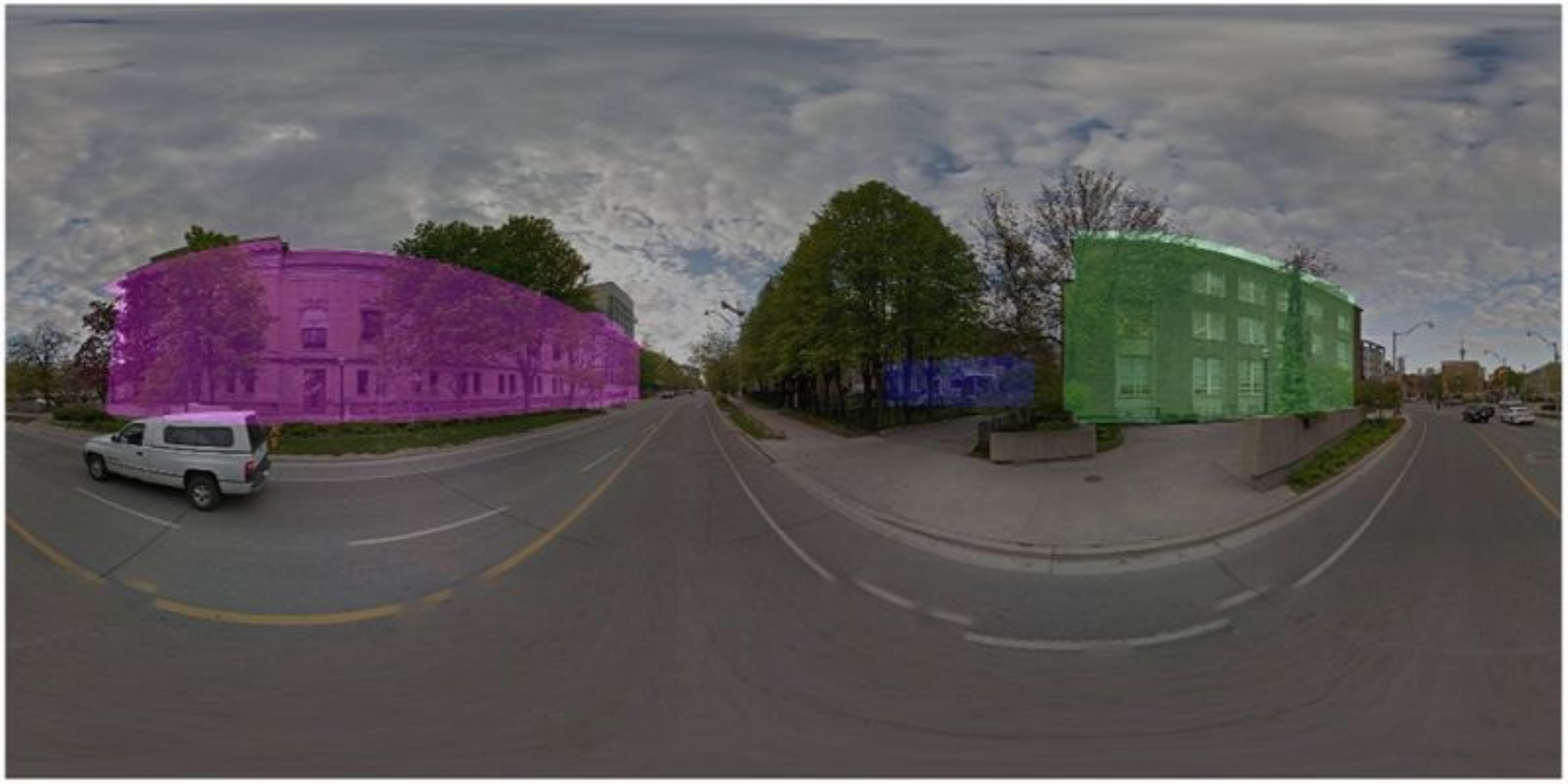}\ 
 \includegraphics[width=0.246\linewidth]{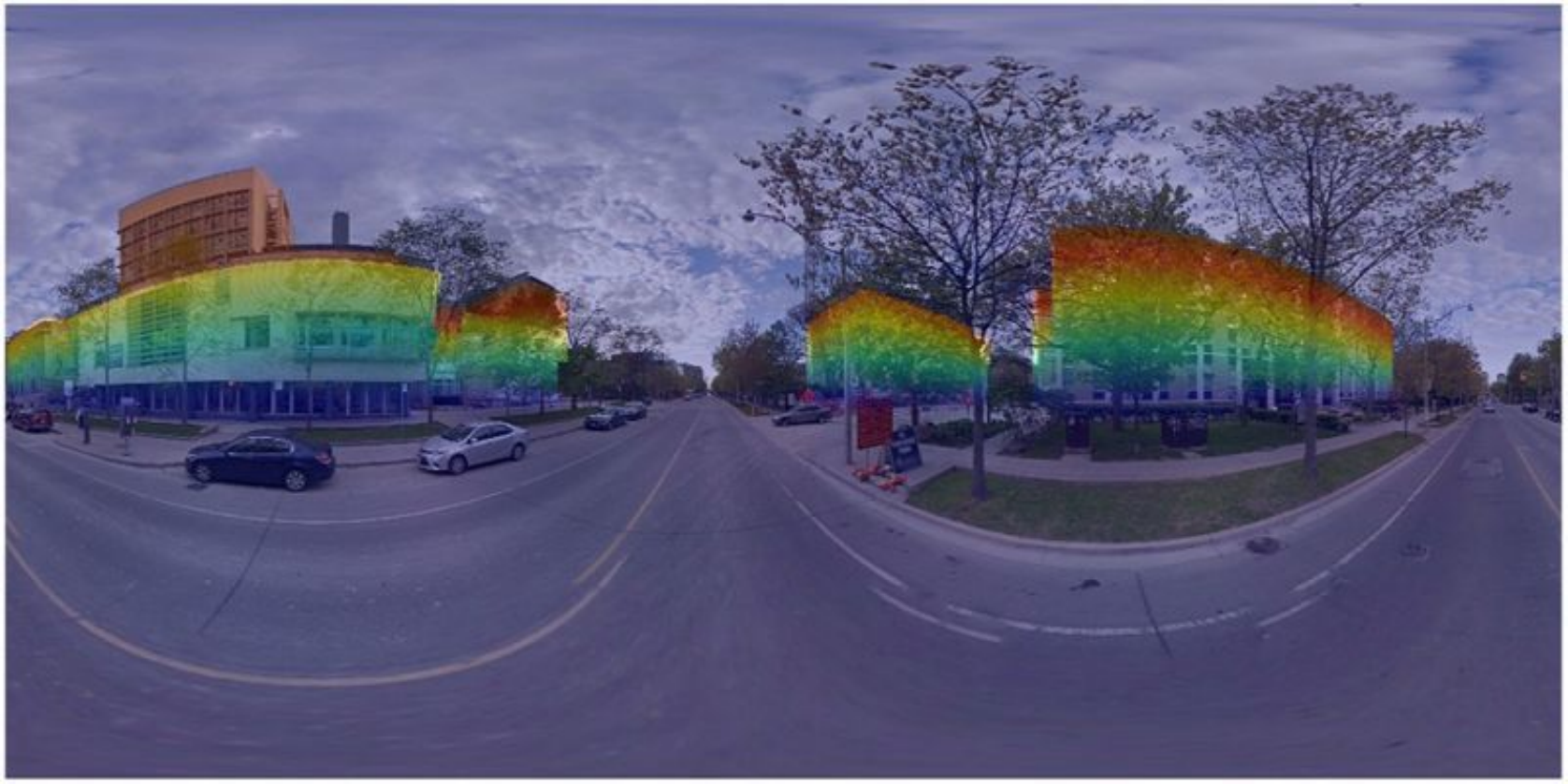}\ 
 \includegraphics[width=0.246\linewidth]{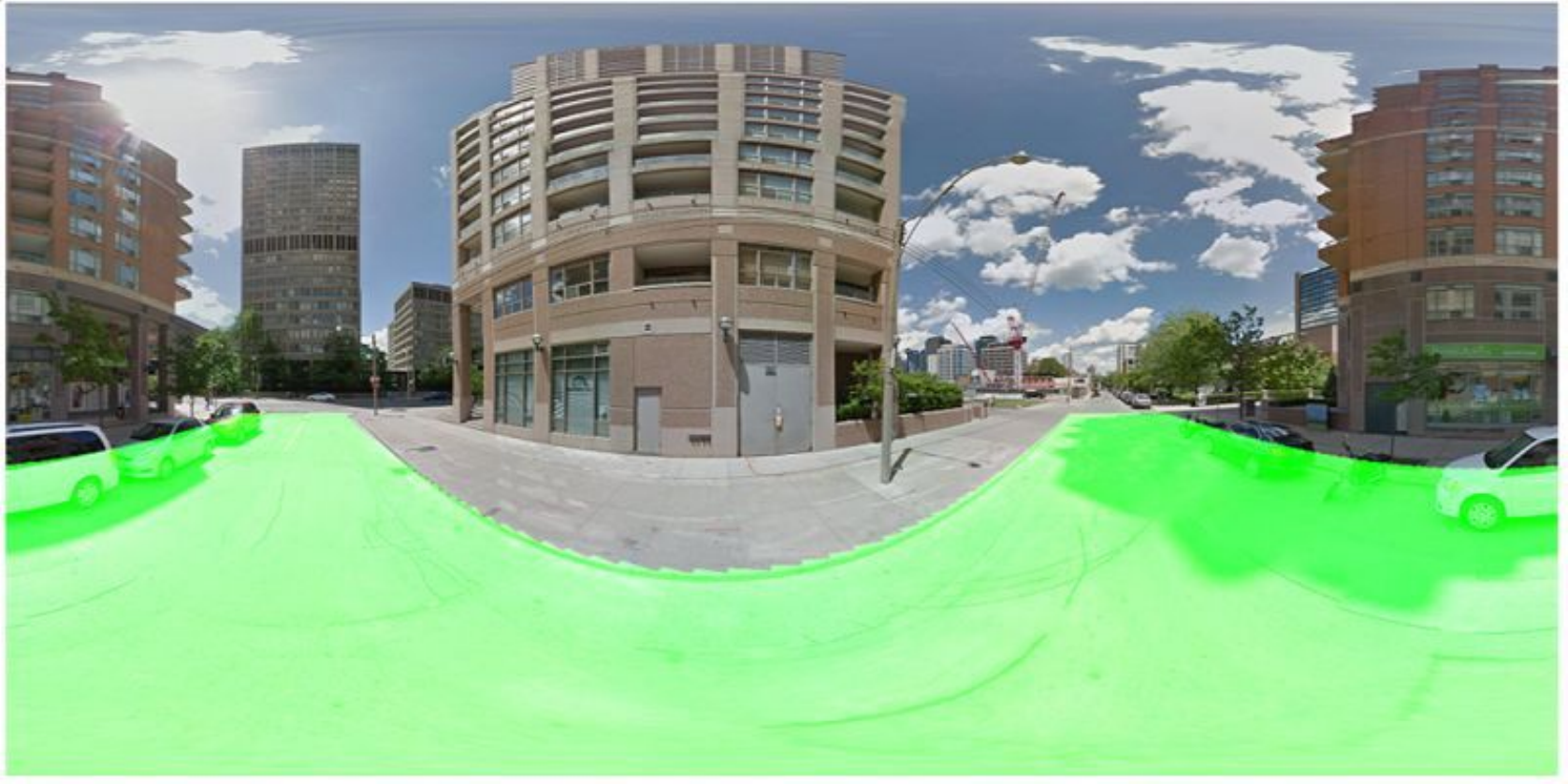}\ 
 \includegraphics[width=0.246\linewidth]{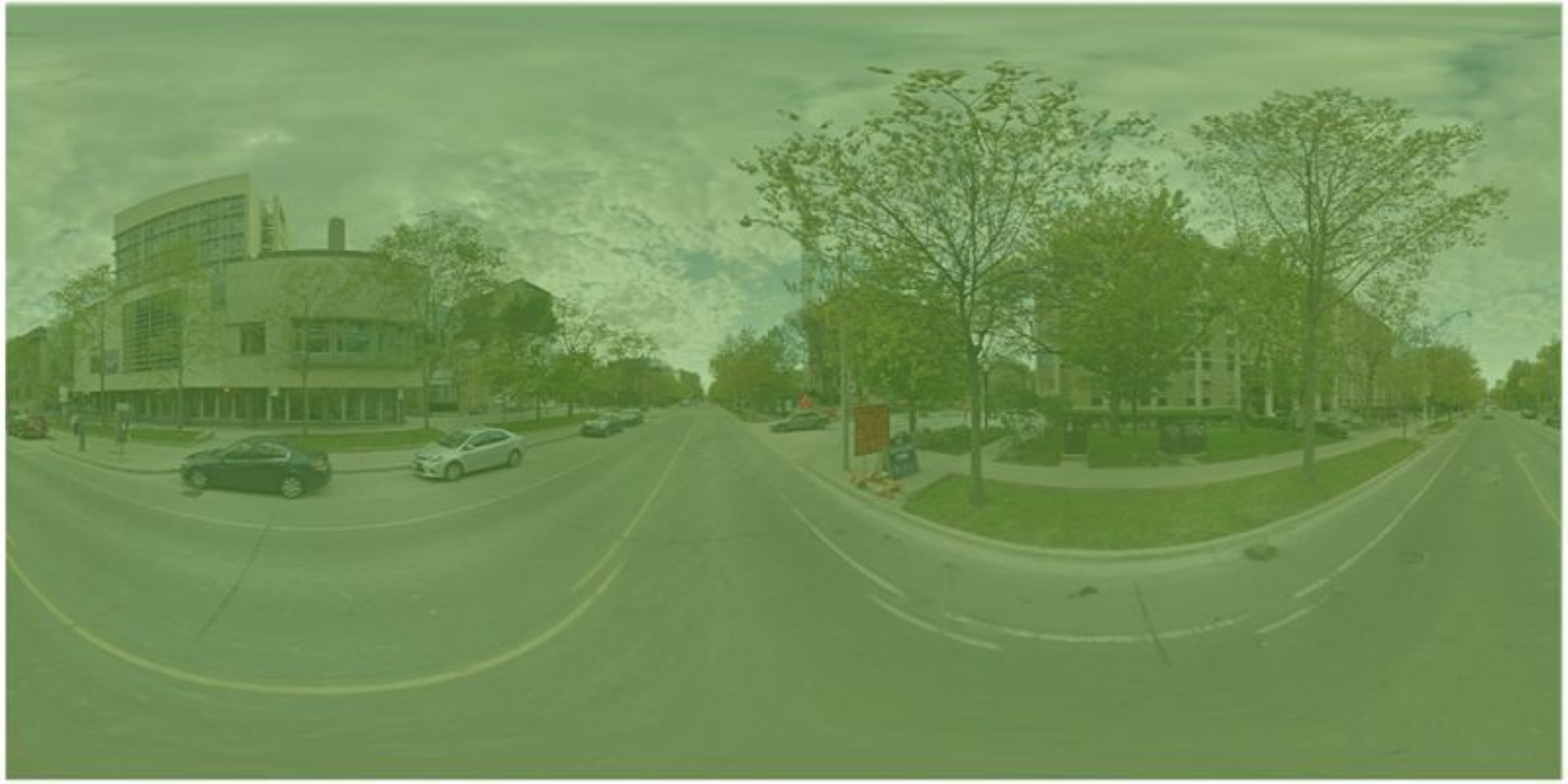}
  \end{subfigure}
 \end{center}
  \end{minipage}
   \vspace{-0.2cm}
 \caption{Summary of the TorontoCity benchmark. Data source: aerial RGB image, streetview panorama, GoPro, stereo image, street-view LIDAR, airborne LIDAR; Maps: buildings and roads, 3D buildings, property meta-data; Tasks: semantic segmentation, building height estimation, instance segmentation, road topology, zoning segmentation and classification. }
 \label{fig:intro}
 \end{figure*}

However, current large scale datasets suffer from two shortcomings. First, they 
have been captured by a small set of sensors with similar perspectives 
of the world, e.g., internet photos for ImageNet or 
cameras/LIDAR mounted on top of a car in the case of KITTI. Second, 
they do not contain rich semantics and 3D information at a large-scale.
We refer the reader to Fig. \ref{fig:dataset} for an analysis of existing datasets. 

In this paper, we argue that the field is in need of large scale benchmarks that allow joint reasoning about geometry, grouping and 
semantics. This has  been commonly referred 
to as the three R's of computer vision. 
Towards this goal, we have created the  
TorontoCity benchmark,  covering the full greater Toronto area (GTA) with $712.5km^2$ of land, $8439km$ of road and around $400,000$ buildings. According to the census, $6.8 million$ people live in the GTA, which is around $20\%$ of the population of Canada.
We have gathered a wide range of views of the city: from the overhead perspective, 
we have aerial images captured during 
four different years as well as  LIDAR from airborne. 
From the ground, we have HD panoramas as well as stereo, Velodyne LIDAR and Go-pro data captured from a moving vehicle driving around in the city. 
We are also augmenting the dataset with a 3D camera as well as  imagery captured from drones. 

Manually labeling such a large scale dataset is  not feasible. Instead,  we propose to utilize  different sources of  high-precision maps  to create our ground truth. 
Compared to online map services such as  OpenStreetMap \cite{osm} and Google Maps, our maps are much more accurate and contain richer meta-data, which we exploit to create a wide variety of diverse benchmarks.  This includes tasks such as building height estimation (reconstruction), road centerline and curb extraction, building instance segmentation,  building contour extraction (reorganization), semantic labeling and scene type classfication (recognition). 
Participants can exploit any subset of the data (e.g., aerial and ground images)
 to solve these tasks.

One of the main challenges in creating TorontoCity was aligning  the maps to all data sources such that the maps can produce accurate ground truth. 
While the aerial data was perfectly aligned
, this is not the case of  the panoramas where 
geolocalization is fairly noisy. 
To alleviate this problem, we have created a set of tools which allow us to reduce the labeling task to a simple verification process, speeding up  labeling, thus making TorontoCity possible. 

We perform a pilot study using the aerial images captured in 2011 as well as the ground panoramas. 
Our experiments show that  state-of-the-art methods work well on some tasks, such as semantic segmentation and scene classification. However, tasks such as instance segmentation, contour extraction and height estimation remain an open challenge. 
We believe our benchmark provides a great platform for developing and evaluating new ideas, particularly  techniques that can leverage different viewpoints of the world. 
We plan to extend the current set of benchmarks in the near future with tasks such as building reconstruction, facade parsing, tree detection and tree species classification as well as traffic lights and traffic sign detection, for which our maps provide accurate ground truth. 
We have only scratched 
the surface of TorontoCity's full potential.  



%% file: related.tex
 \section{Related Work}


 \begin{figure*}[t]
  \vspace{-0.7cm}
 \centering
\begin{subfigure}[t]{0.26\textwidth}
 \includegraphics[width=\textwidth]{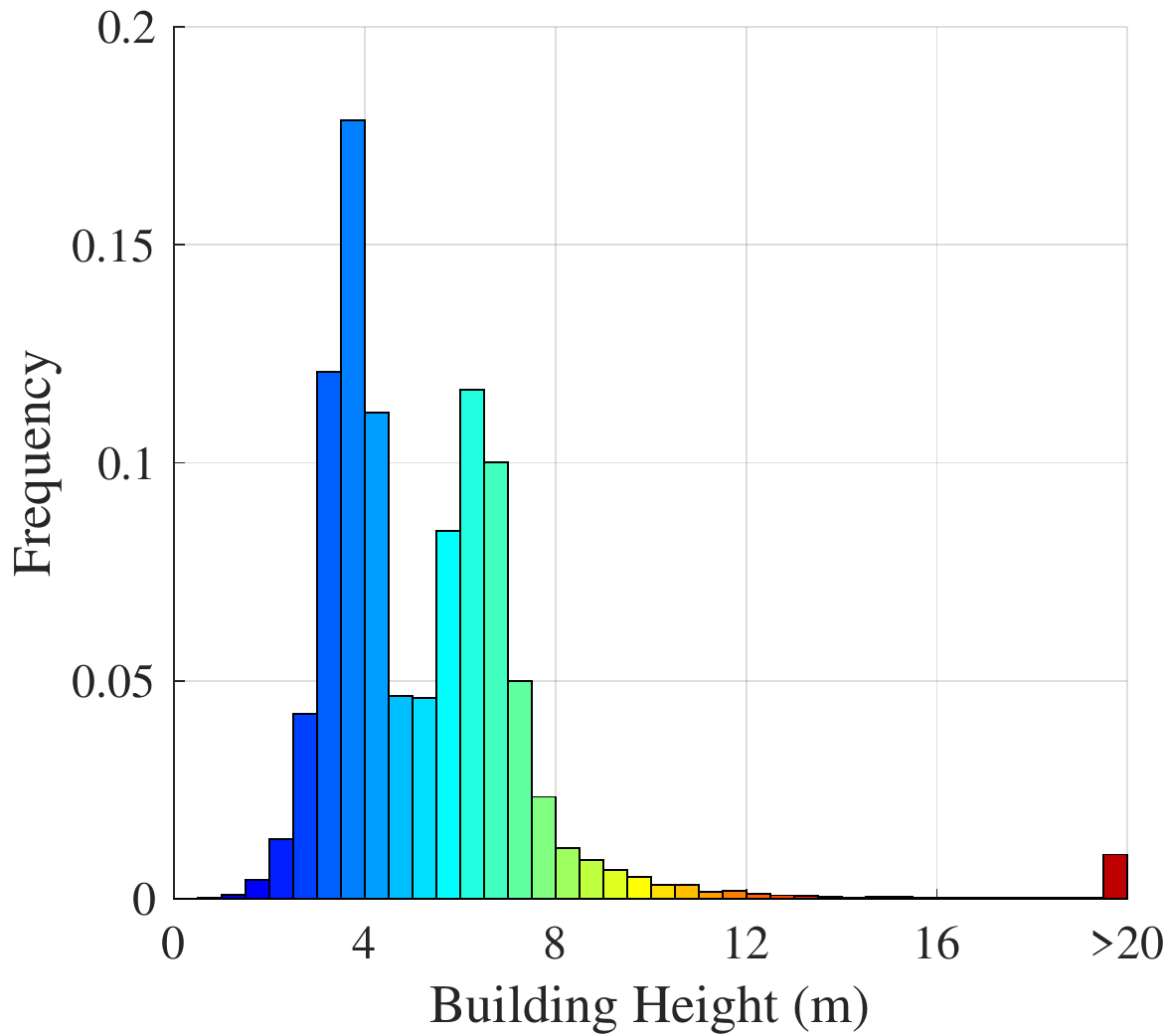}
  \end{subfigure}
\begin{subfigure}[t]{0.26\textwidth}
 \includegraphics[width=\textwidth]{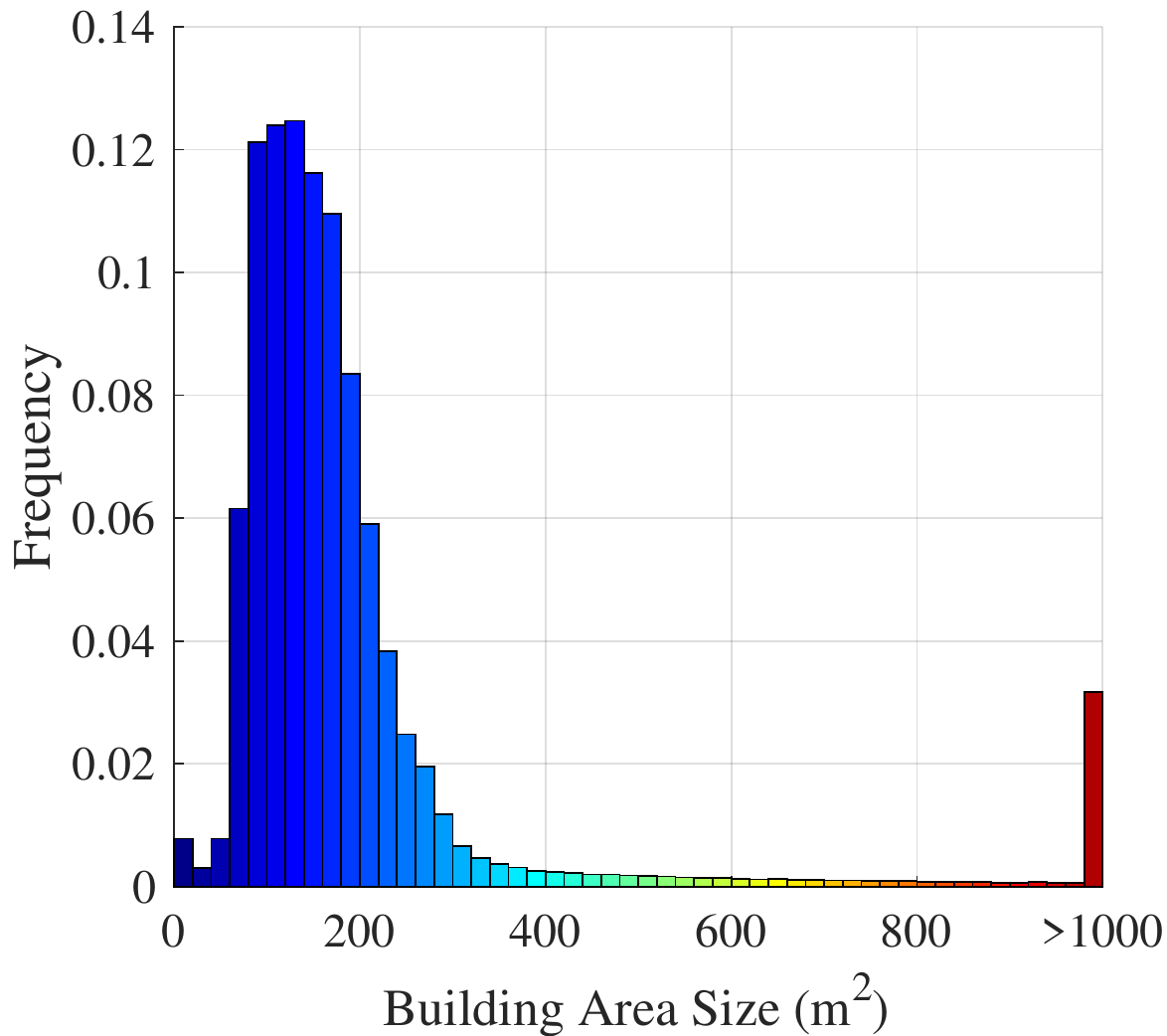}
 \end{subfigure}
 \begin{subfigure}[t]{0.3\textwidth}
 \includegraphics[width=\textwidth]{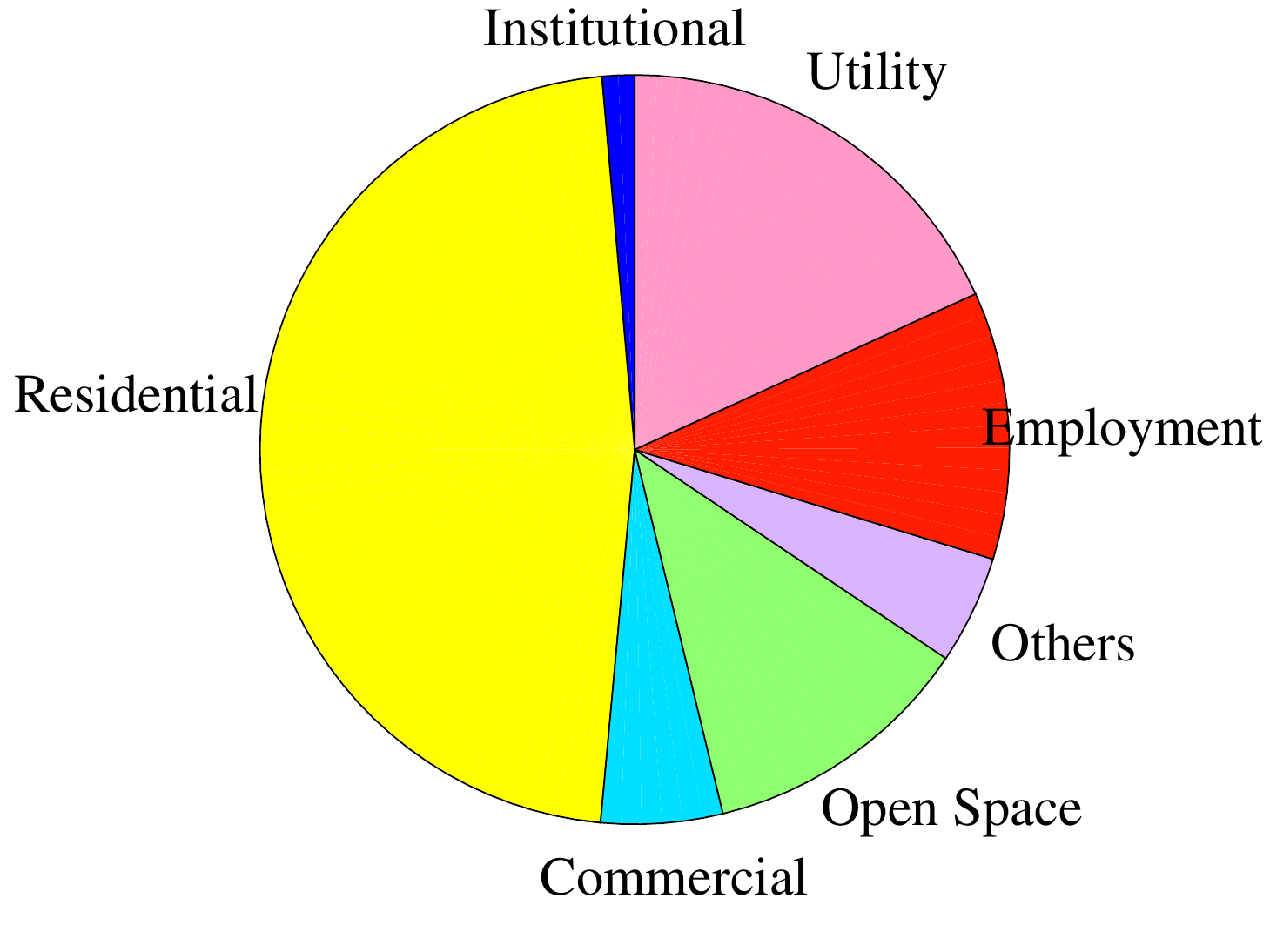}
 \end{subfigure}
\begin{center}
\footnotesize
\begin{tabular}{c|ccccccc}
Dataset & ISPRS & TUM-DLR & Aerial KITTI & KITTI & RobotCar & Ours \\ \hline
Location & Vaihingen/Toronto & Munich & Karlsruhe & Karlsruhe & Oxford & Toronto  \\
Aerial Coverage ( $km^2$) & 3.49+1.45 & 8.32 & 3.23 & -  & - &  712 \\
Ground Coverage ($km$) & - & $<$1 & $<$20 & 39.2 & 10 & $>$1000(pano)\footnote{We are keeping collecting more data} \\
Aerial RGB & yes & yes & yes & - & - &  yes \\
Drone RGB & - & yes & - & - & - &  yes \\
Aerial LIDAR & yes & yes & - & - & - &  yes \\
Ground Perspective & - & - & - & yes & yes &  yes \\
Ground Panorama & - & - & - & - & yes &  yes \\
Ground Stereo & - & yes & - & yes & yes & yes \\
Ground LIDAR & - & yes & - & yes & yes &  yes \\
Aerial Resolution (pixel/$cm^2$) & 8 & 50 & 9 & - & - & 10 \\
Repeats & - & - & - & partial & x10 & x4 (aerial) \\
Top Semantic GT (\# of classes) & 100\% (8) & - & 100\% (4) & - & - & 100\% (2 + 8) \\
Top Geometric GT (source) & dense (lidar) & dense (lidar) & - & - & - & dense (map+lidar) \\
Ground Semantic GT (\# of classes) & - & - & dense (4) & object (3) & - & dense (2) / image (6) \\
Ground Geometric GT (source) & - & - & - & sparse (lidar) & sparse (lidar) & dense (map+lidar) \\
\hline
\end{tabular}
 \end{center}
 \vspace{-0.5cm}
 \caption{Statistics of our data and comparison of current state-of-the-art urban benchmarks and datasets.}
  \label{fig:dataset}
 \end{figure*}

Automatic mapping, reconstruction and semantic labeling from urban scenes have been an important topic for many decades. Several benchmarks have been proposed to tackle subsets of these tasks.
KITTI  \cite{kitti} is composed of  stereo images and LIDAR data collected from a moving vehicle, and evaluates SLAM, optical flow, stereo and road segmentation tasks. Cityscapes \cite{cityscape} focuses on semantic and instance annotations of images captured from a car. Aerial-KITTI \cite{a-kitti} augments the KITTI dataset with aerial imagery of a subset of Karlsruhe to encourage  reasoning of semantics from both ground and bird's eye view.

The photometry community has  developed several benchmarks towards urban scene understanding \cite{tum-dlr, isprs-multi-platform, isprs-urban, isprs-2d, oxfordcar}. TUM-DLR \cite{tum-dlr} and ISPRS Multi-Platform \cite{isprs-multi-platform} benchmarks contain imagery captured   through multiple perspectives from UAV, satellite images and handheld cameras. Oxford RobotCar contains lidar point cloud and stereo images captured from a vehicle \cite{oxfordcar}. However, these benchmarks do not offer any semantic ground-truth for benchmarking purposes.
Perhaps the most closely related dataset to ours is the ISPRS Urban classification and building reconstruction benchmark \cite{isprs-urban}, where the task is to extract urban object, such as building, road and trees from both aerial images and airborne laserscanner point clouds.
However, this dataset 
has a relatively small coverage and does not provide ground-view imagery.
In contrast, TorontoCity is more than two orders of magnitude bigger. 
 Furthermore, we offer many different perspectives through various sensors, along with diverse semantic and geometric benchmarks with accurate ground-truth. 
 {The readers may refer to Fig.~\ref{fig:dataset} for a detailed comparison against previous datasets. }

A popular alternative is to use synthetic data to generate large scale benchmarks
 \cite{middlebury, sintel, flyingthings, synthia, halflife, scenenet, deepdriving, play}.
 Through 3D synthetic scenes and photo-realistic renderers 
 large-scale datasets can be easily created. To date, however, %
 these datasets have been focused on a single view of the world. This contrasts TorontoCity.
 Unlike other 
 benchmarks, our input  is real-world imagery, and the large-scale 3D models are a high-fidelity modeling of the real world rather than a synthetic scene.

Maps have been proven useful for many computer vision and robotics applications \cite{kitti3d, nyc3dcar, hdmap, registree, a-kitti}, including  vehicle detection and pose estimation \cite{nyc3dcar},  
semantic labeling and monocular depth estimation \cite{kitti3d} as well as HD-map extraction 
\cite{a-kitti}.
However, there has been a lack of literature that exploit 
maps as ground-truth to build  benchmarks. This is mainly due to both the lack of high-fidelity maps to provide pixel-level annotation and the lack of accurately georeferenced imagery that aligned well with the maps. One exception is  \cite{registree}, where the streetree catalog  is used to generate ground-truth for tree detection.
\cite{3dmatching} utilizes 3D building models to generate correspondences  from multiple streetview images. In this paper, we use maps to create
multiple 
benchmarks for reconstruction, recognition and reorganization from many different views of the world.

%
%
 \begin{figure}[t]
  \vspace{-0.3cm}
 \centering

 \includegraphics[width=0.48\linewidth]{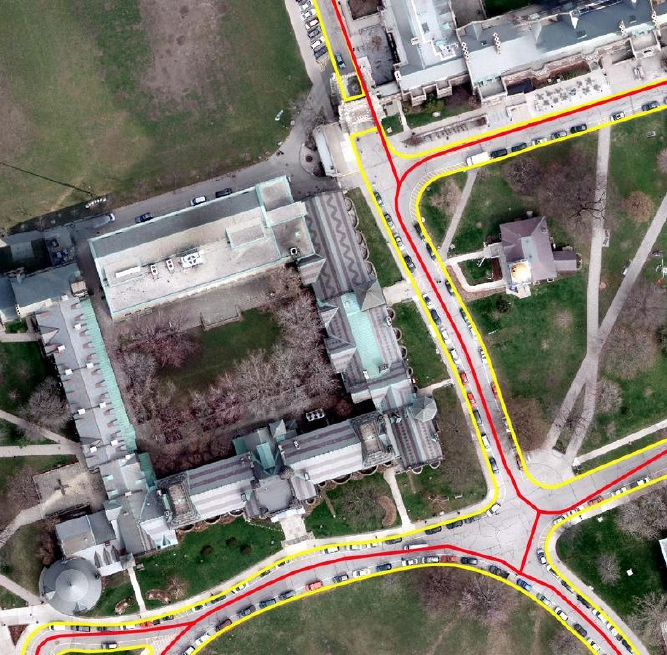}
 \includegraphics[width=0.48\linewidth]{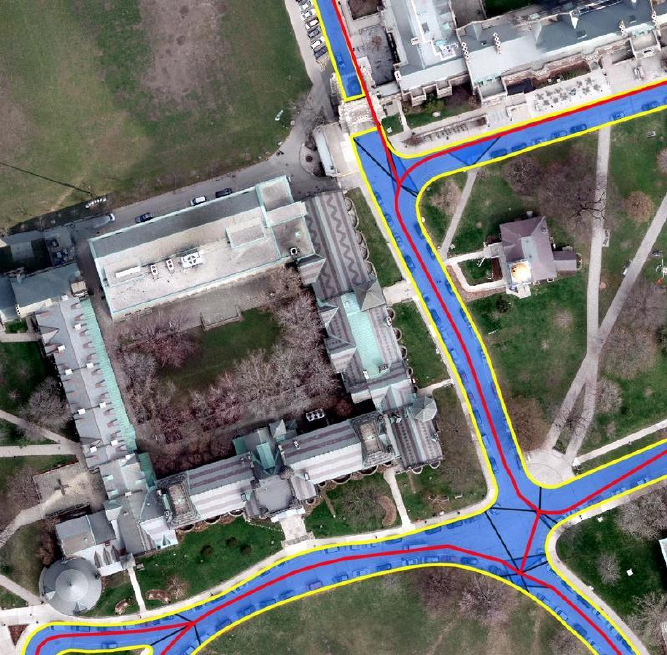}
  \vspace{-0.2cm}
 \caption{Road surface generation: (left) input data with  curbs (yellow) and  center lines (red). Extracted road surface is the union of polygons shown in blue  and black. Note that a formulation ensuring connectivity is needed, otherwise the road surface would contain holes at intersections.}
 \label{fig:roadSurfaceGeneration}
    \vspace{-3mm}
 \end{figure}

 \begin{figure*}[t]
 \vspace{-.7cm}
 \centering
\begin{subfigure}[t]{0.36\textwidth}
 \includegraphics[width=\textwidth]{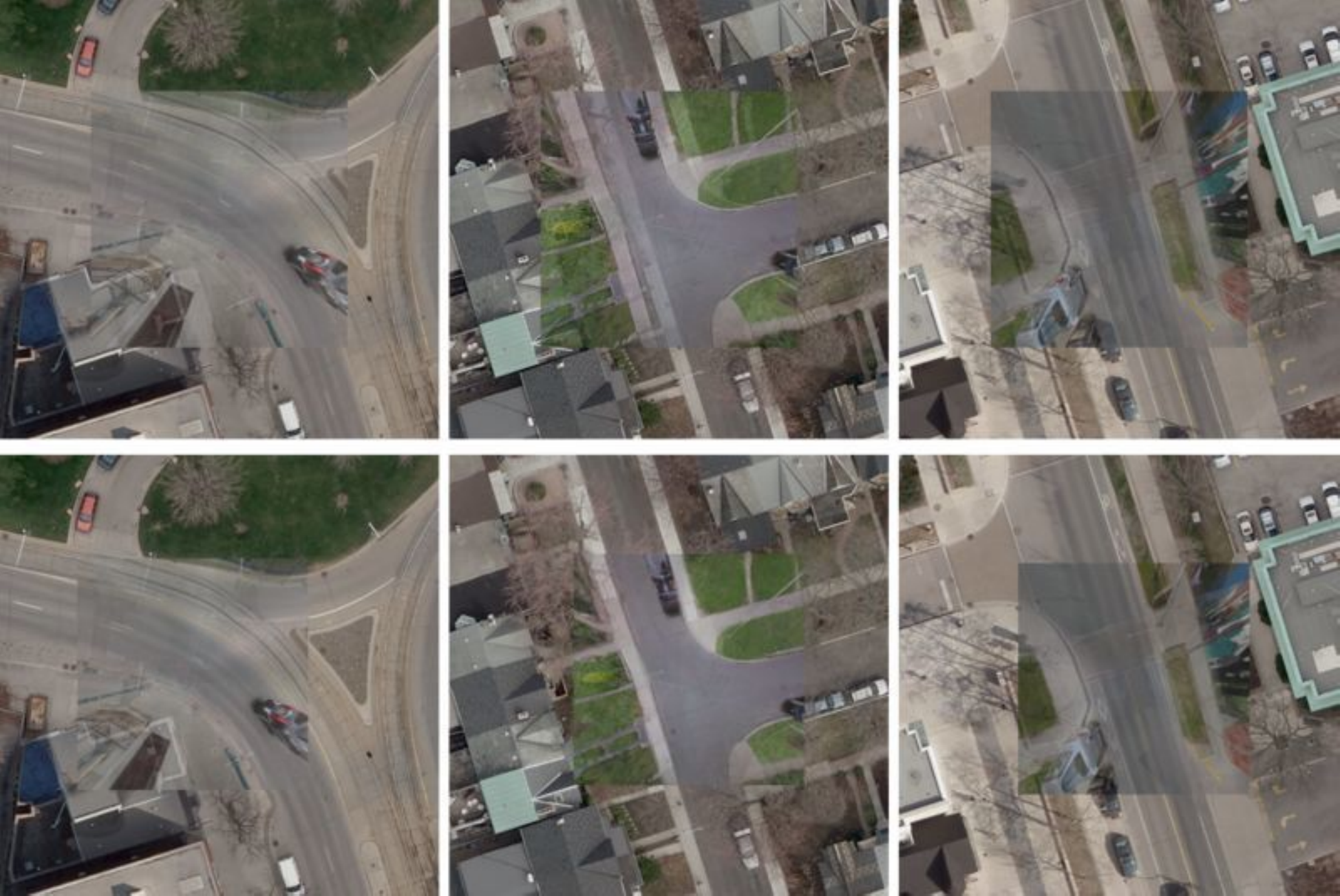}
 \caption{NCC: before vs. after }
 \end{subfigure}
 \begin{subfigure}[t]{0.25\textwidth}
 \includegraphics[width=0.95\textwidth]{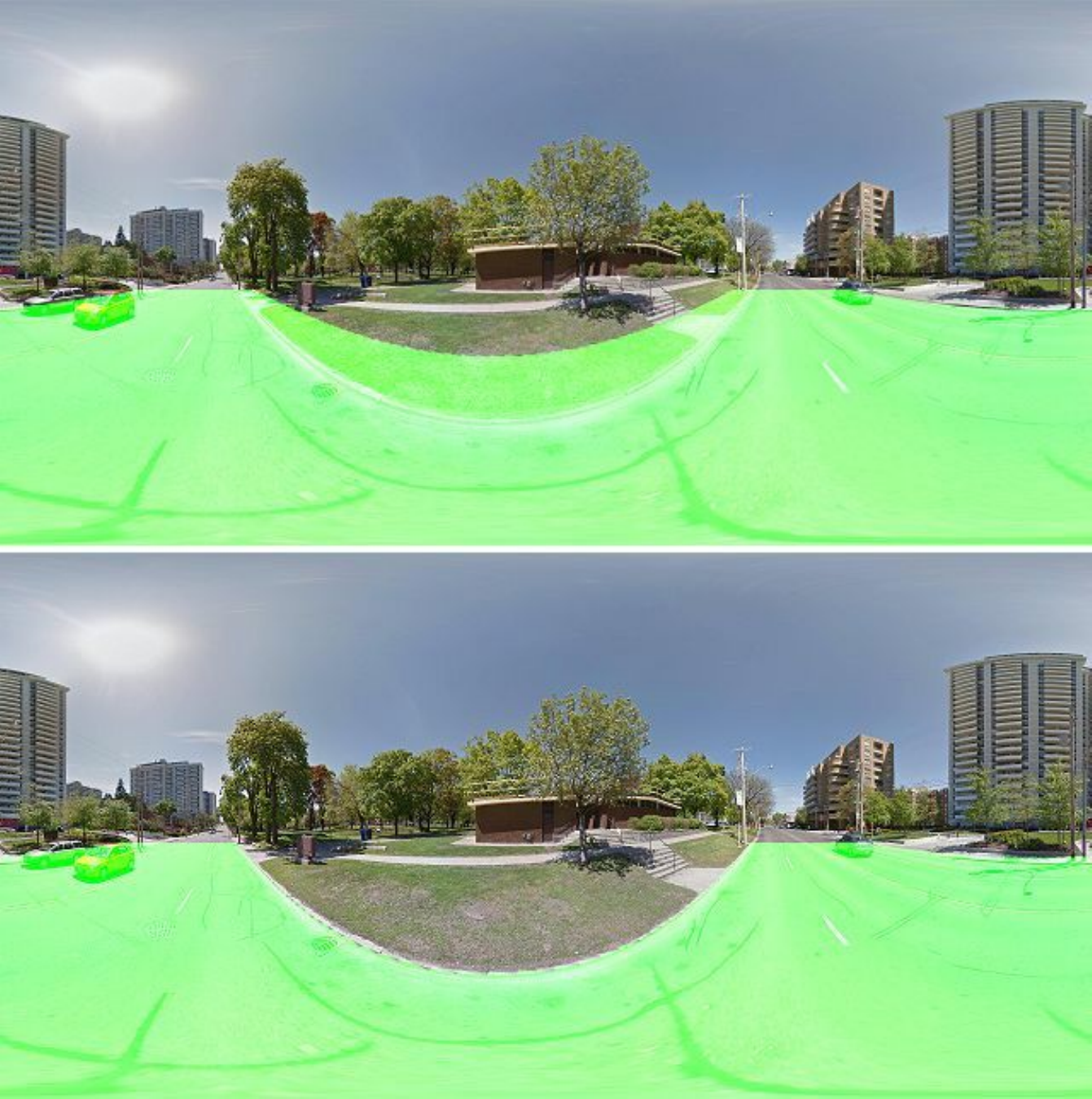}
 \caption{Overlay: before vs. after}
 \end{subfigure}
 \begin{subfigure}[t]{0.28\textwidth}
 \includegraphics[width=0.98\textwidth]{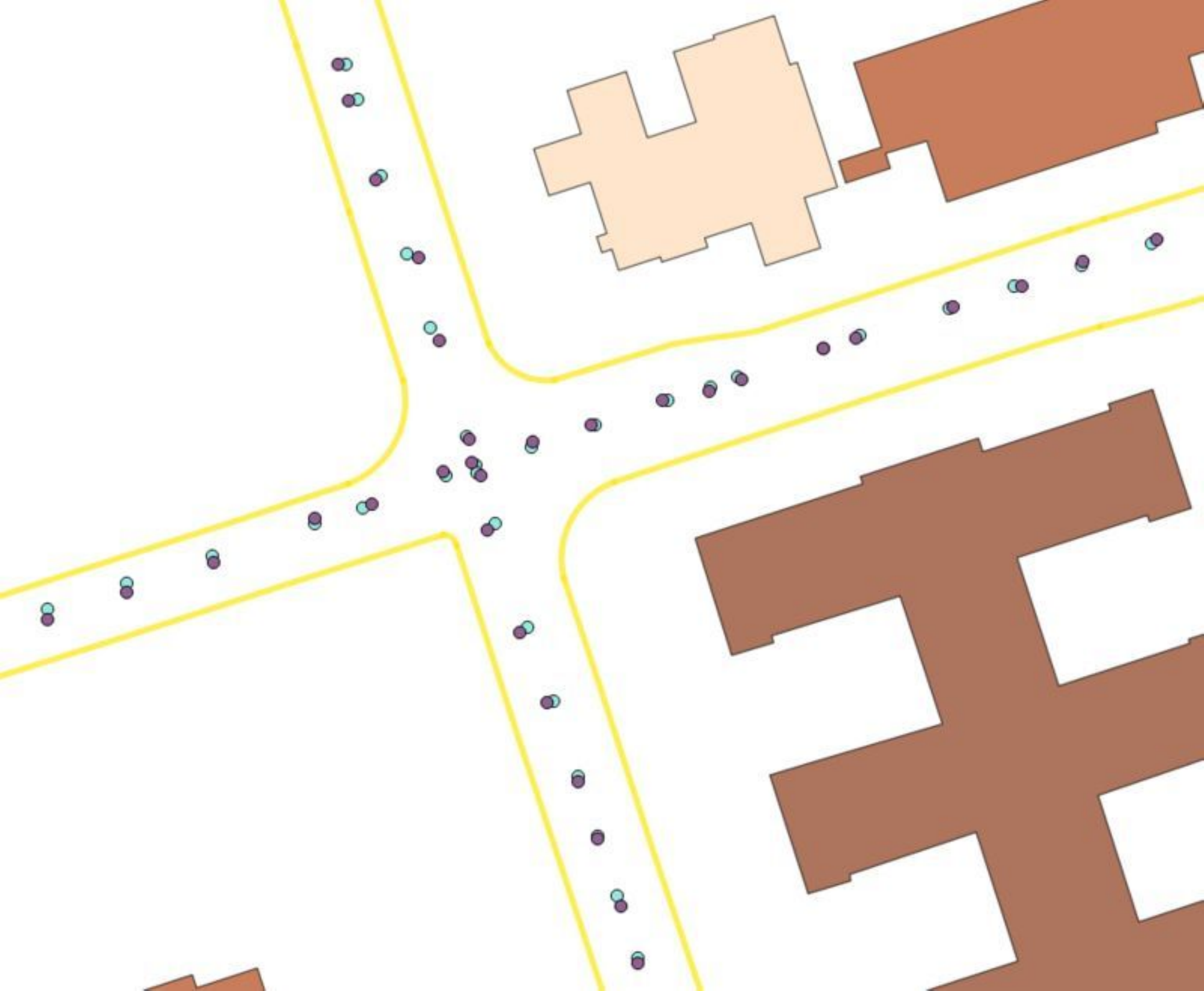}
  \caption{Location: {\color{cyan}before} vs. {\color{violet}after}}
  \end{subfigure}
   \vspace{-3mm}
 \caption{Ground-aerial alignment  }
 \label{fig:ncc}
   \vspace{-2mm}
 \end{figure*}

%% file: dataset.tex
 \begin{figure*}[t]
 \centering
 \begin{subfigure}[t]{0.16\textwidth}
  \adjincludegraphics[width=\textwidth,trim={0 0 {.5\width}  {.6\width}},clip]{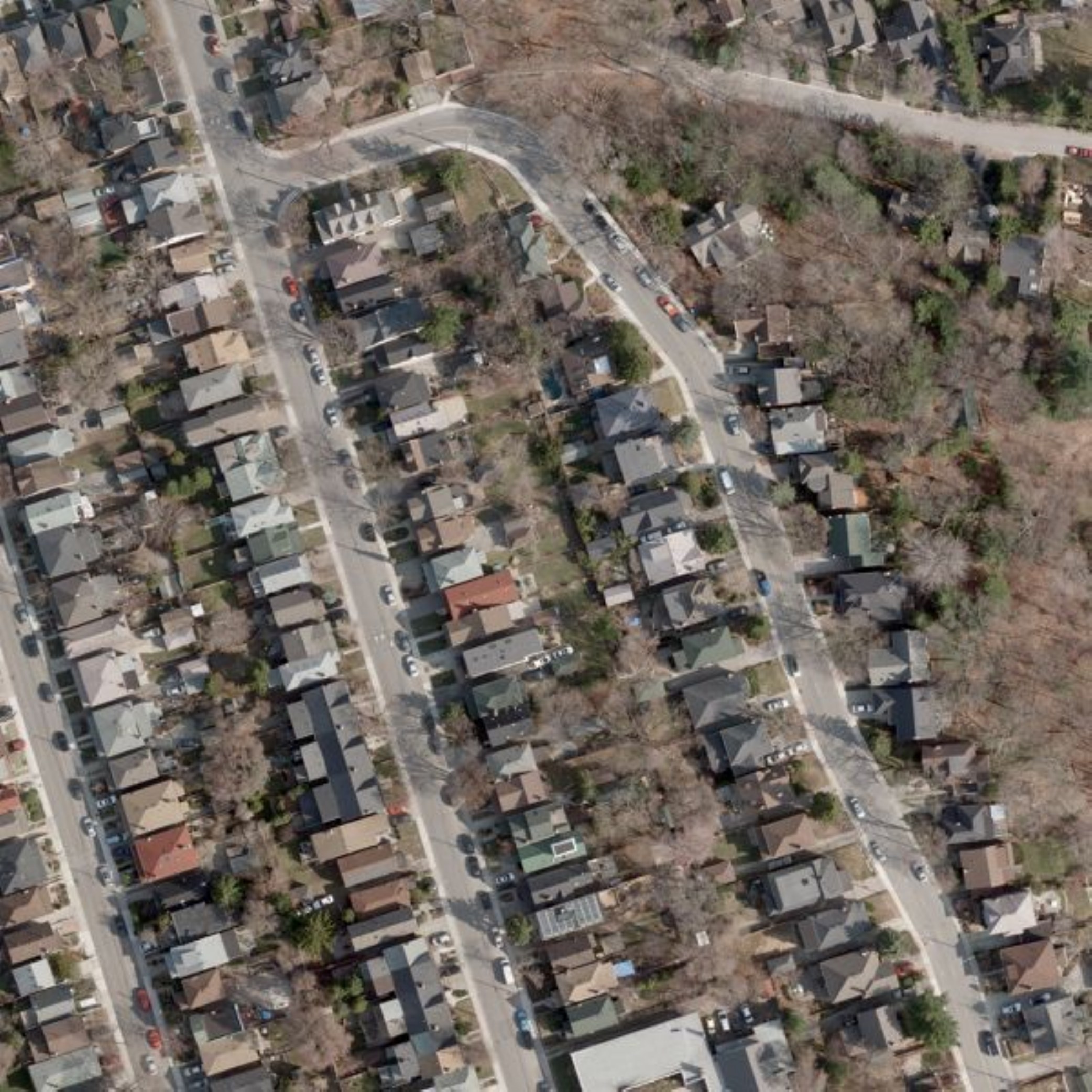} 
 
  \vspace{0.5mm}
  
  \adjincludegraphics[width=\textwidth,trim={0 {.2\width} 0 0},clip]{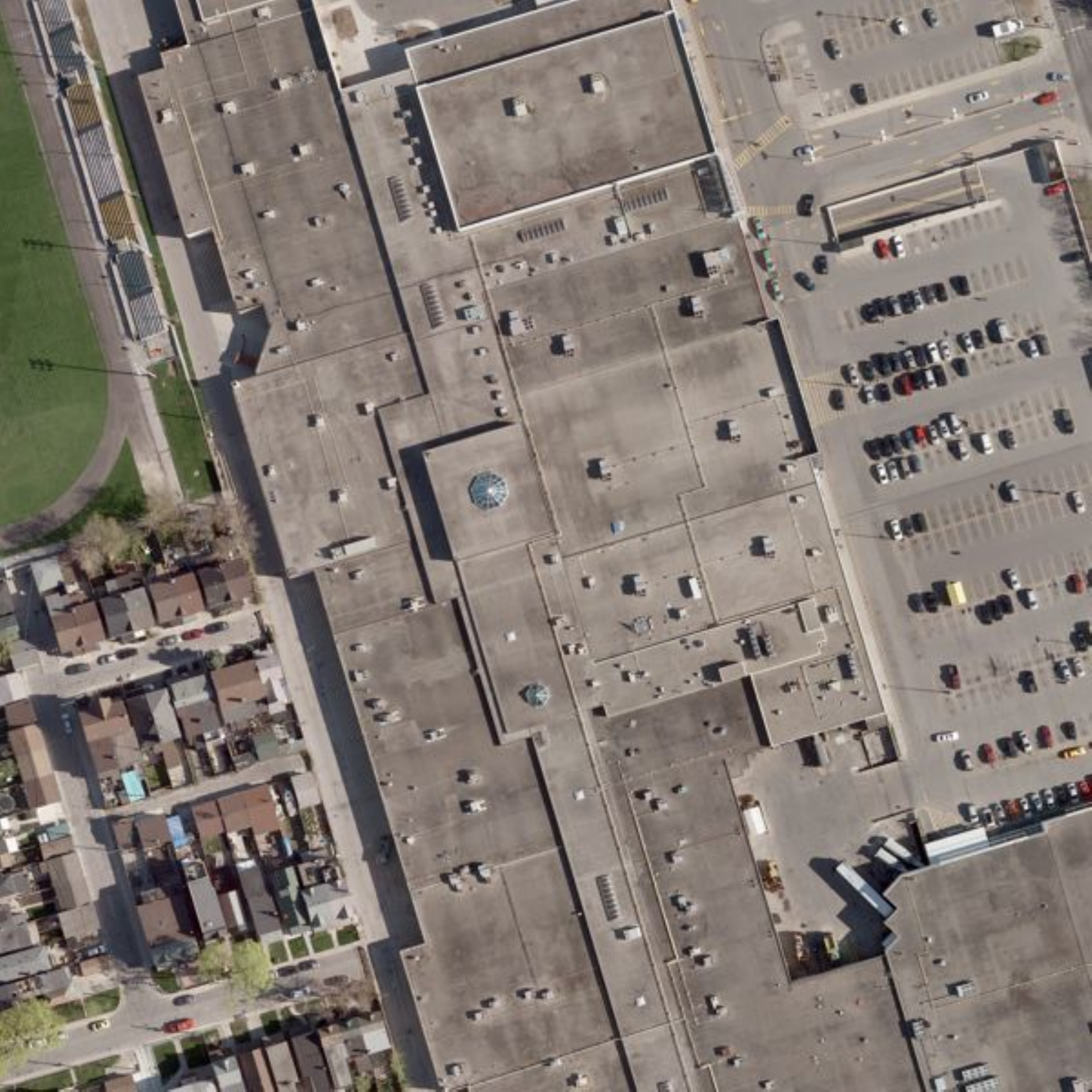} 
 \caption{Input}
  \end{subfigure}
\begin{subfigure}[t]{0.16\textwidth}
  \adjincludegraphics[width=\textwidth,trim={0 0 {.5\width}   {.6\width}},clip]{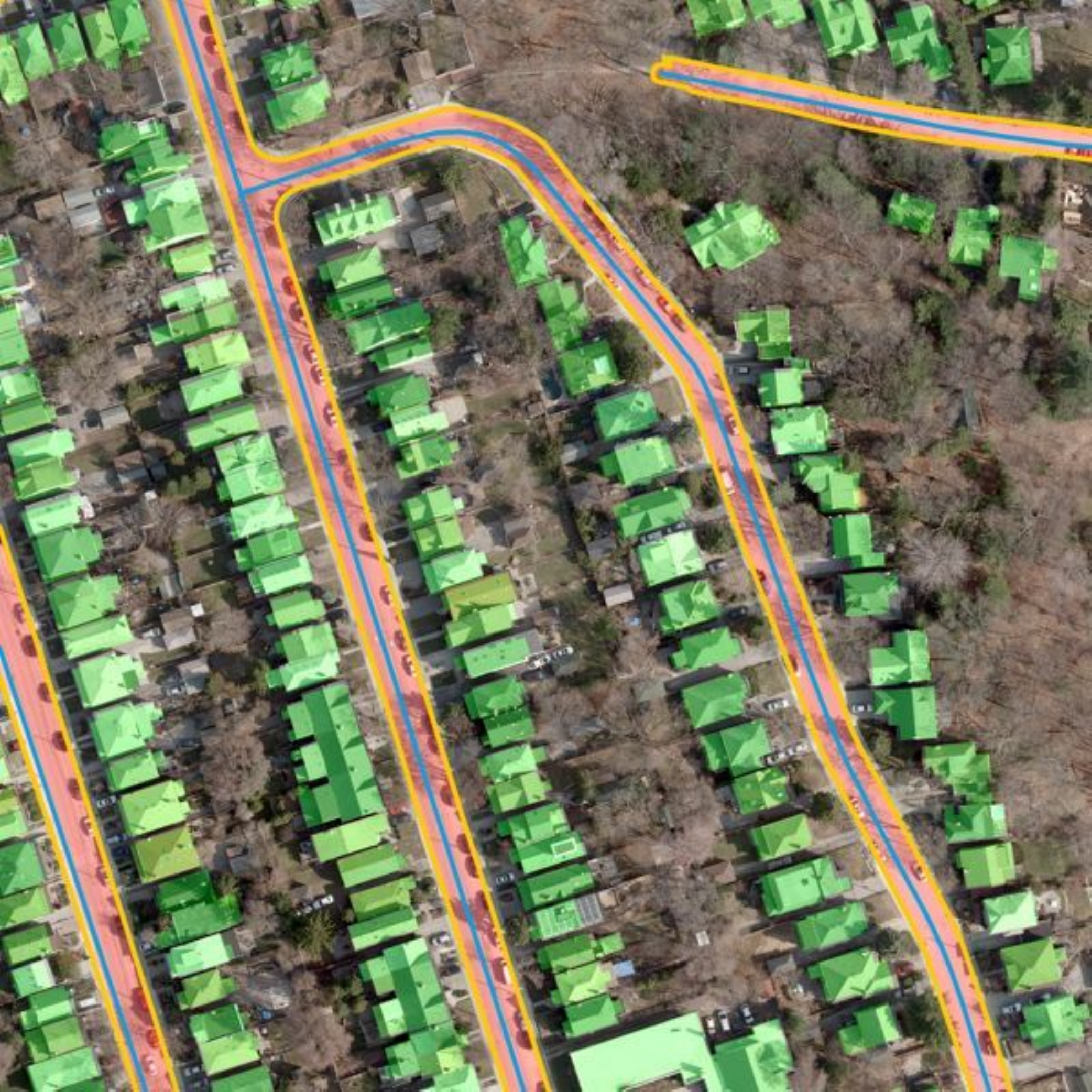}
 
  \vspace{0.5mm}
  
   \adjincludegraphics[width=\textwidth,trim={0 {.2\width} 0 0},clip]{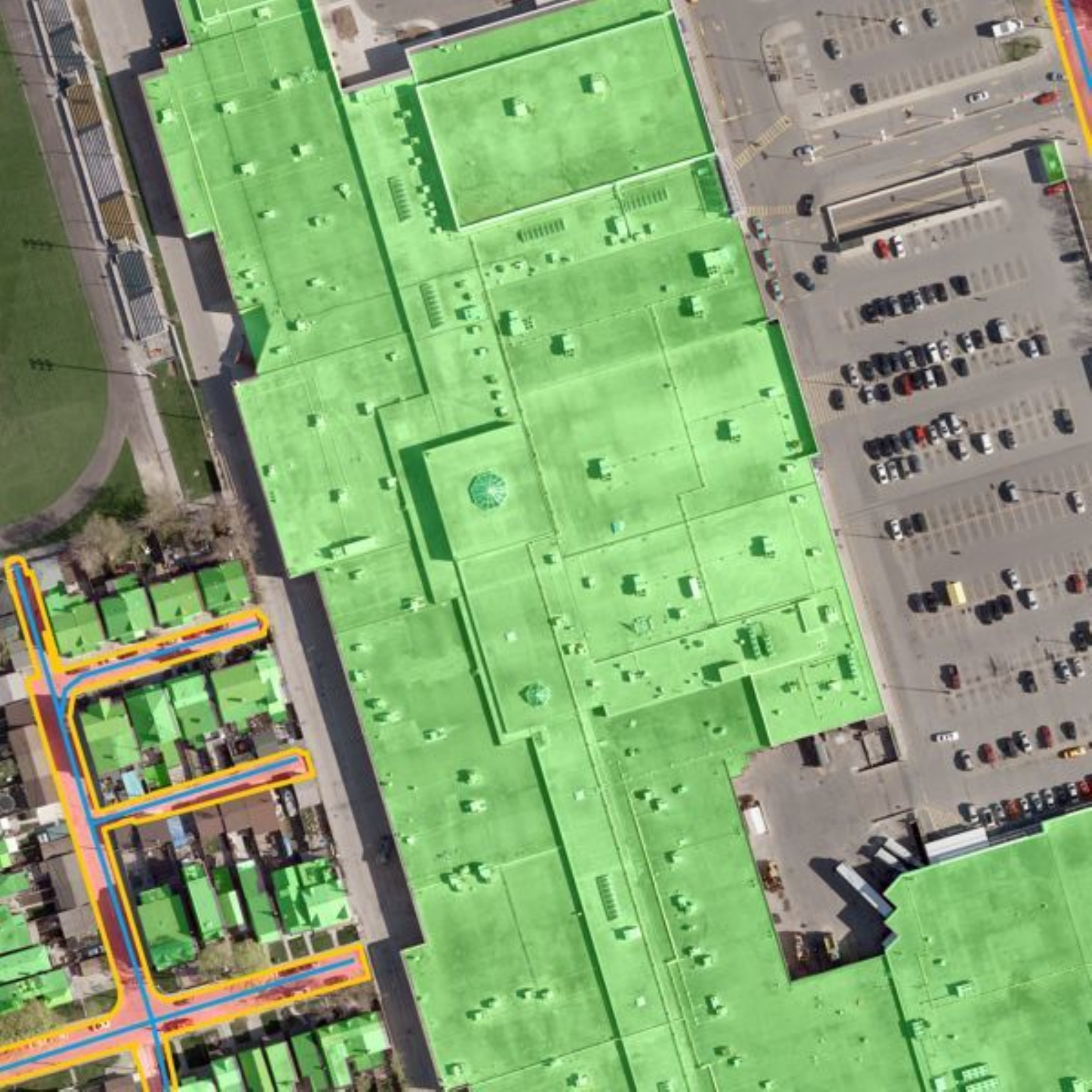} 
 \caption{GT}
 \end{subfigure}
 \begin{subfigure}[t]{0.16\textwidth}
  \adjincludegraphics[width=\textwidth,trim={0 0 {.5\width}   {.6\width}},clip]{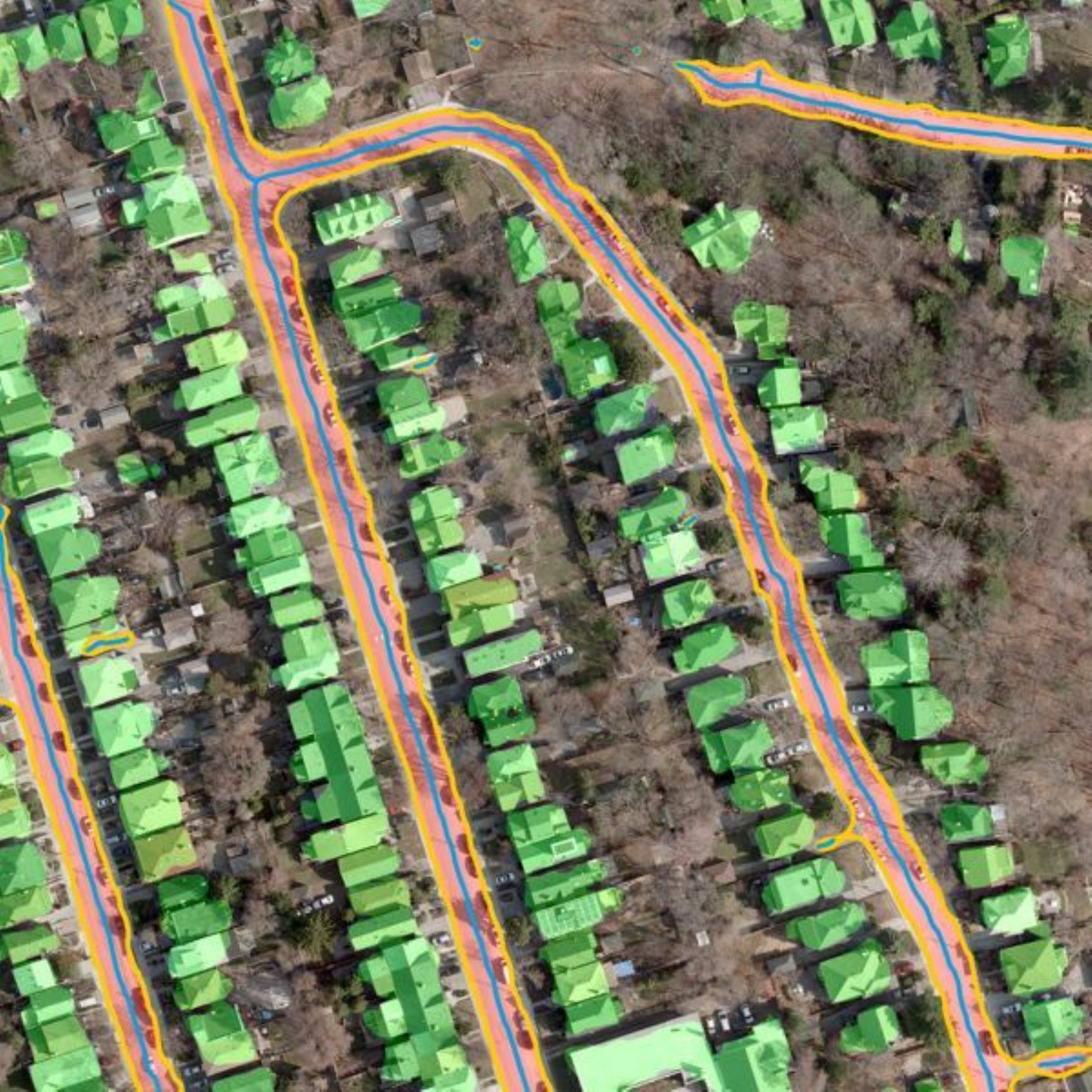} 
 
  \vspace{0.5mm}
  
   \adjincludegraphics[width=\textwidth,trim={0 {.2\width} 0 0},clip]{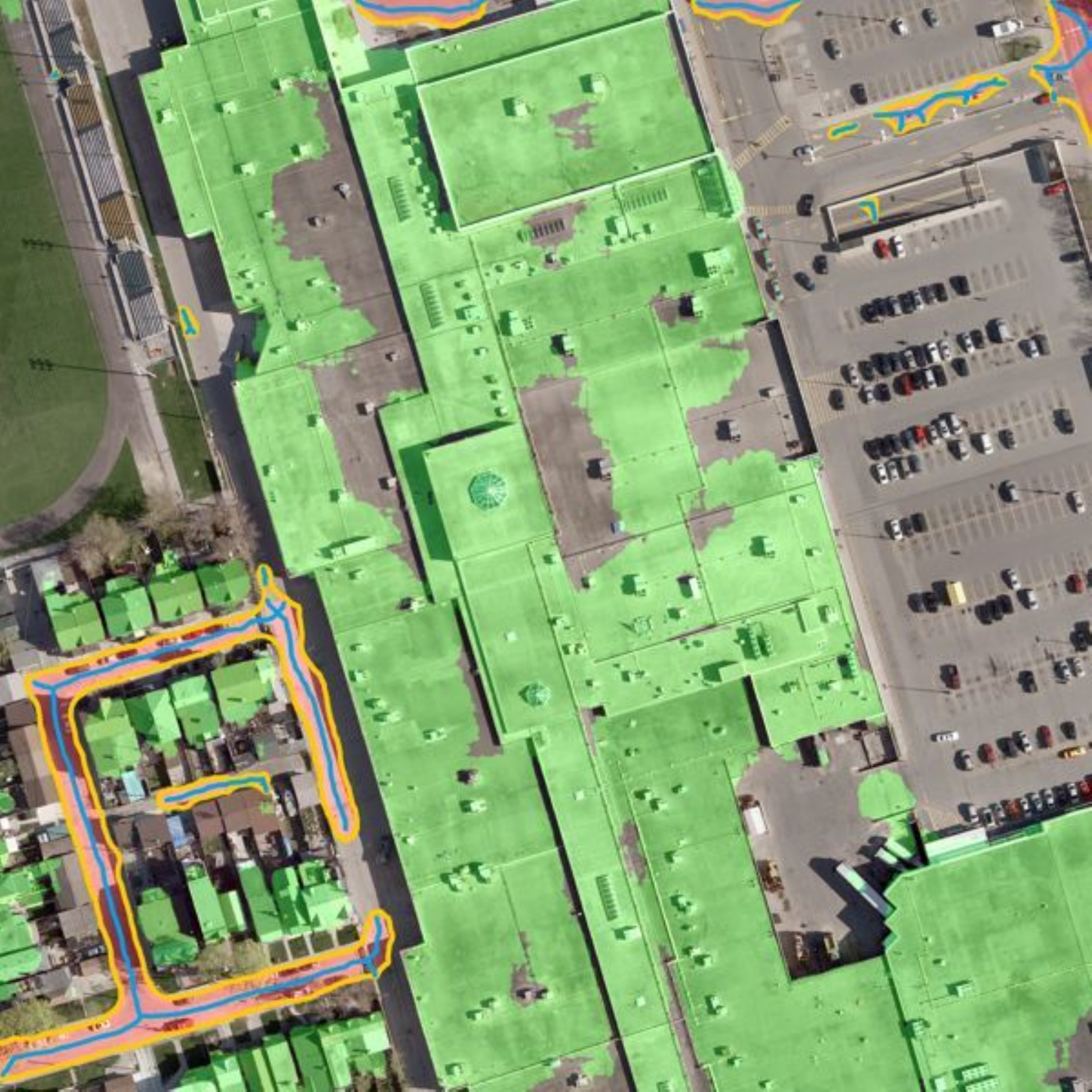}
 \caption{ResNet56}
 \end{subfigure}
\begin{subfigure}[t]{0.16\textwidth}
 \adjincludegraphics[width=\textwidth,trim={0 {.2\width} 0 0},clip]{fig/final_viz/623000_4833500.pdf} 
 
  \vspace{0.5mm}
  
  \adjincludegraphics[width=\textwidth,trim={0 {.2\width} 0 0},clip]{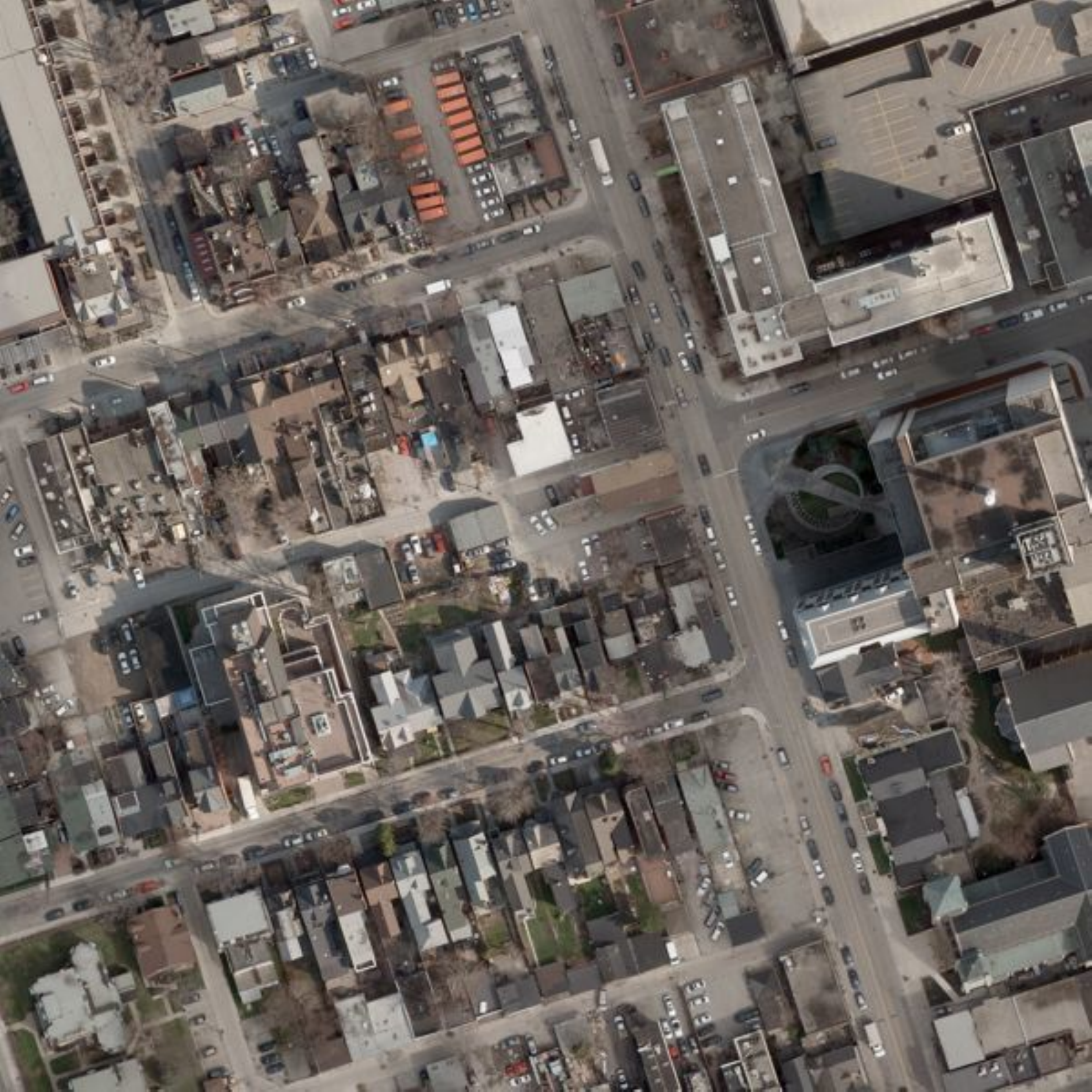}
 \caption{Input}
  \end{subfigure}
\begin{subfigure}[t]{0.16\textwidth}
 \adjincludegraphics[width=\textwidth,trim={0 {.2\width} 0 0},clip]{fig/final_viz/623000_4833500_gt.pdf}
 
  \vspace{0.5mm}
  
  \adjincludegraphics[width=\textwidth,trim={0 {.2\width} 0 0},clip]{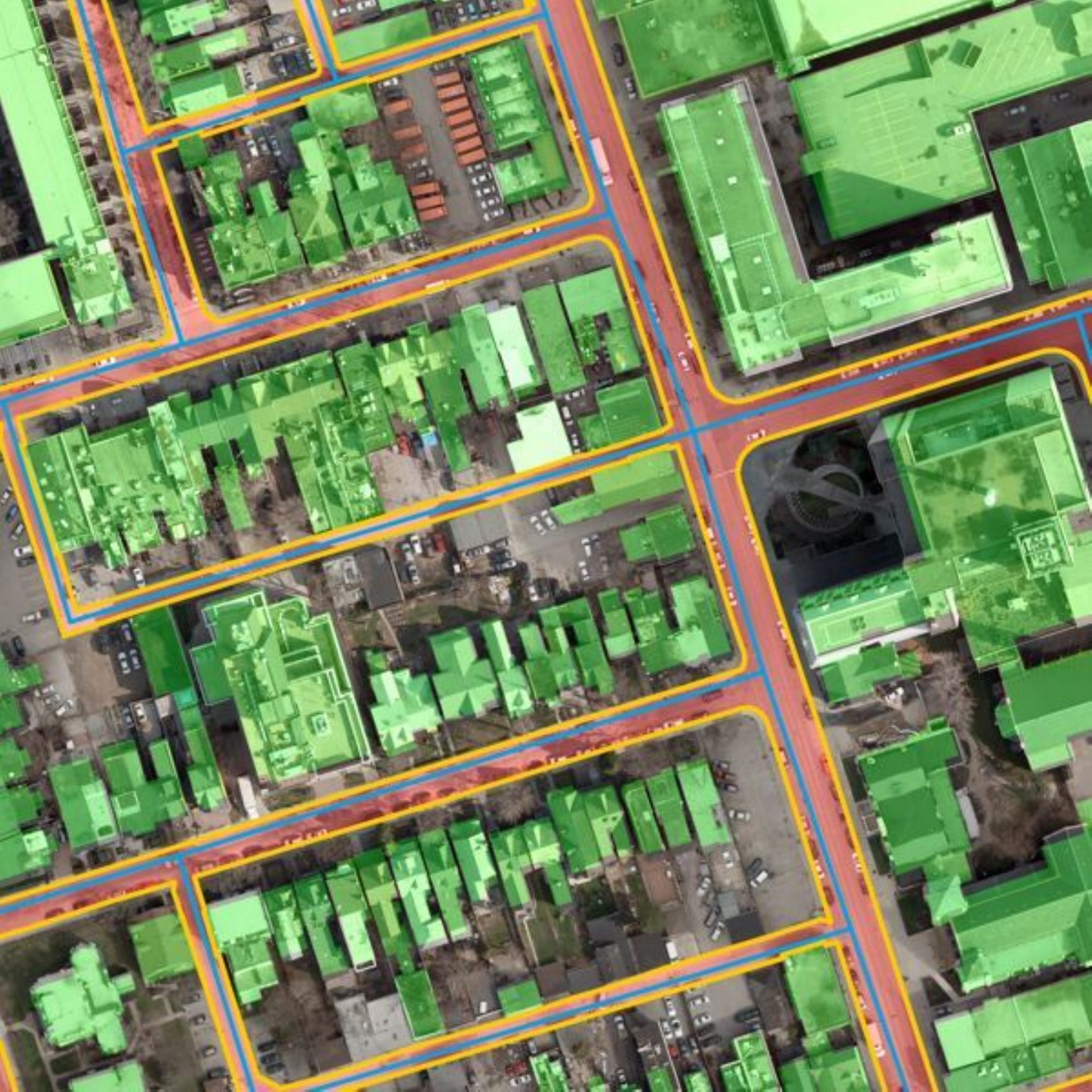} 
 \caption{GT}
 \end{subfigure}
 \begin{subfigure}[t]{0.16\textwidth}
 \adjincludegraphics[width=\textwidth,trim={0 {.2\width} 0 0},clip]{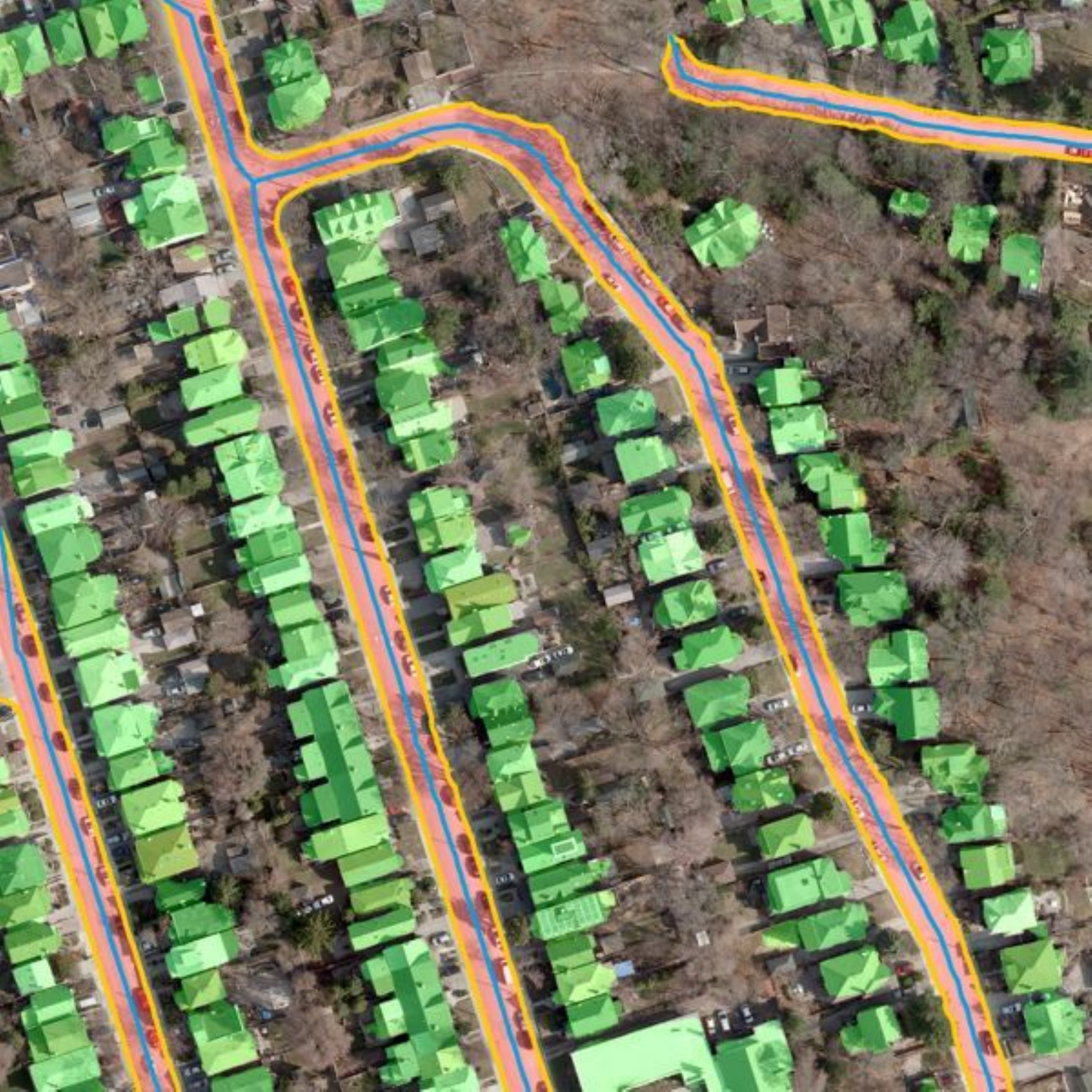} 
 
  \vspace{0.5mm}
  
  \adjincludegraphics[width=\textwidth,trim={0 {.2\width} 0 0},clip]{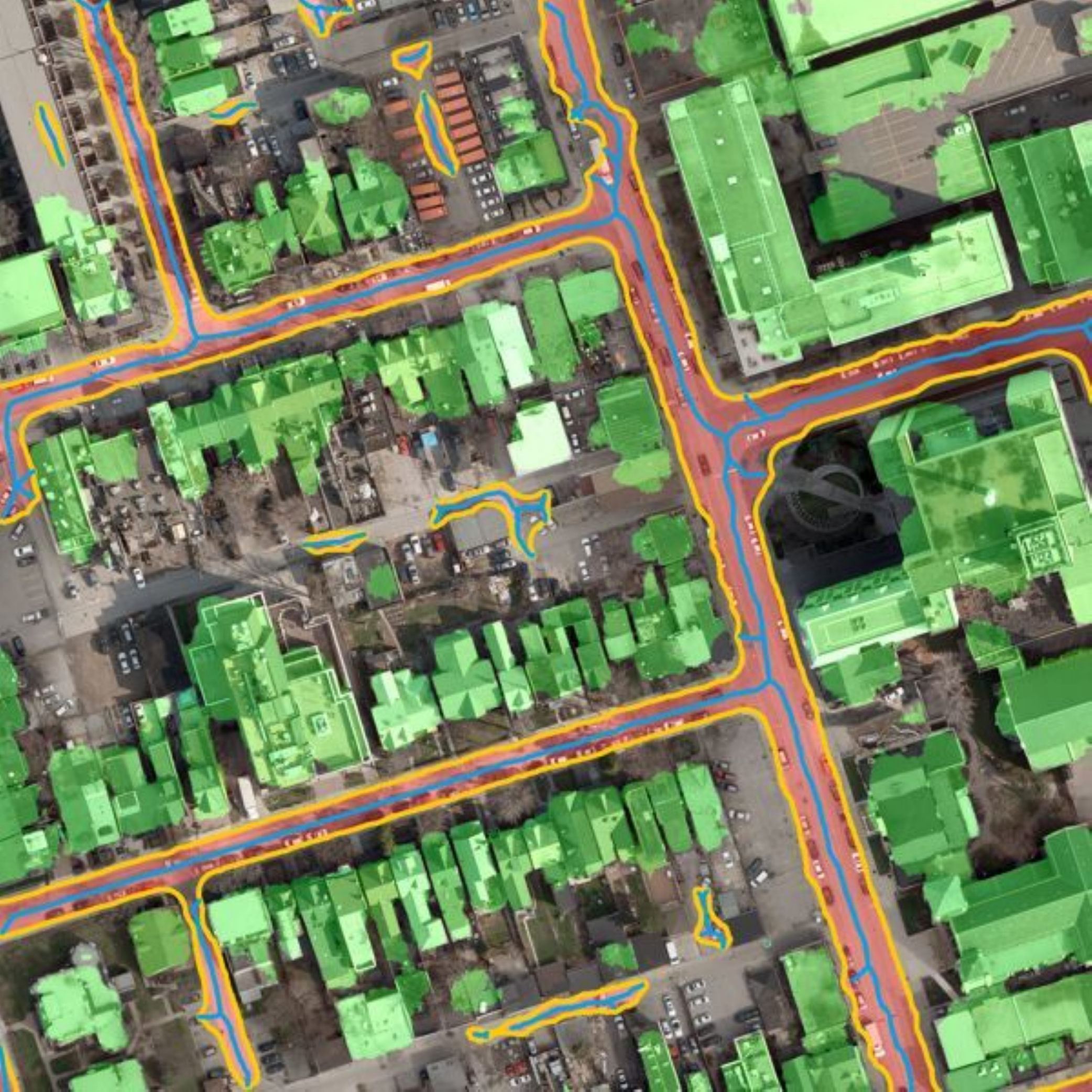} 
 \caption{ResNet56}
 \end{subfigure}
 
%
%
%
%
%
%
    \vspace{-3mm}
 \caption{Examples of aerial semantic segmentation, road curb extraction, and road centerline estimation.}
 \label{fig:aerial-seg}
    \vspace{-2mm}
 \end{figure*}

%
%
%
%

\section{TorontoCity at a Glimpse} 

TorontoCity is an extremely large dataset enabling work on 
many exciting new tasks. 
We first    describe the data 
in detail.  
In the next section we describe our efforts to simplify 
the labeling task, as otherwise it is infeasible to create such a large-scale dataset. We then show  the challenges and metrics that will compose the benchmark. 
Finally, we perform a pilot study of  how current algorithms perform on \shenlong{most tasks}
, and analyze the remaining challenges. 





 \subsection{Dataset}
 

Toronto  is the largest city in Canada, and the fourth largest 
in North America. 
The TorontoCity dataset covers the   greater Toronto area (GTA), 
which contains  $712.5km^2$ of land, $8439km$ of road and around $400,000$ buildings. According to the census $6.8 million$ people live in the GTA, which is around $20\%$ of the population of Canada. 

We have gathered a wide range of views of the city: from the overhead perspective, 
we have aerial images captured during four different years (containing several  seasons) as well as airborne LIDAR.   
From the ground perspective,
we have HD panoramas as well as stereo, Velodyne LIDAR and Go-pro data captured from a moving vehicle driving around the city. 
 In addition, we are augmenting the dataset with a 3D camera as well as  imagery captured from drones. Fig. \ref{fig:intro} depicts some of the data sources that compose our dataset. 
 We now describe the data  in more details and refer the reader to Fig.~\ref{fig:dataset} for a comparison against existing datasets.
 
  \begin{figure*}[t]
  \vspace{-0.5cm}
 \centering
 \begin{subfigure}[t]{0.16\textwidth}
  
  \adjincludegraphics[width=\textwidth,trim={ {.5\width} {.6\width} 0 0},clip]{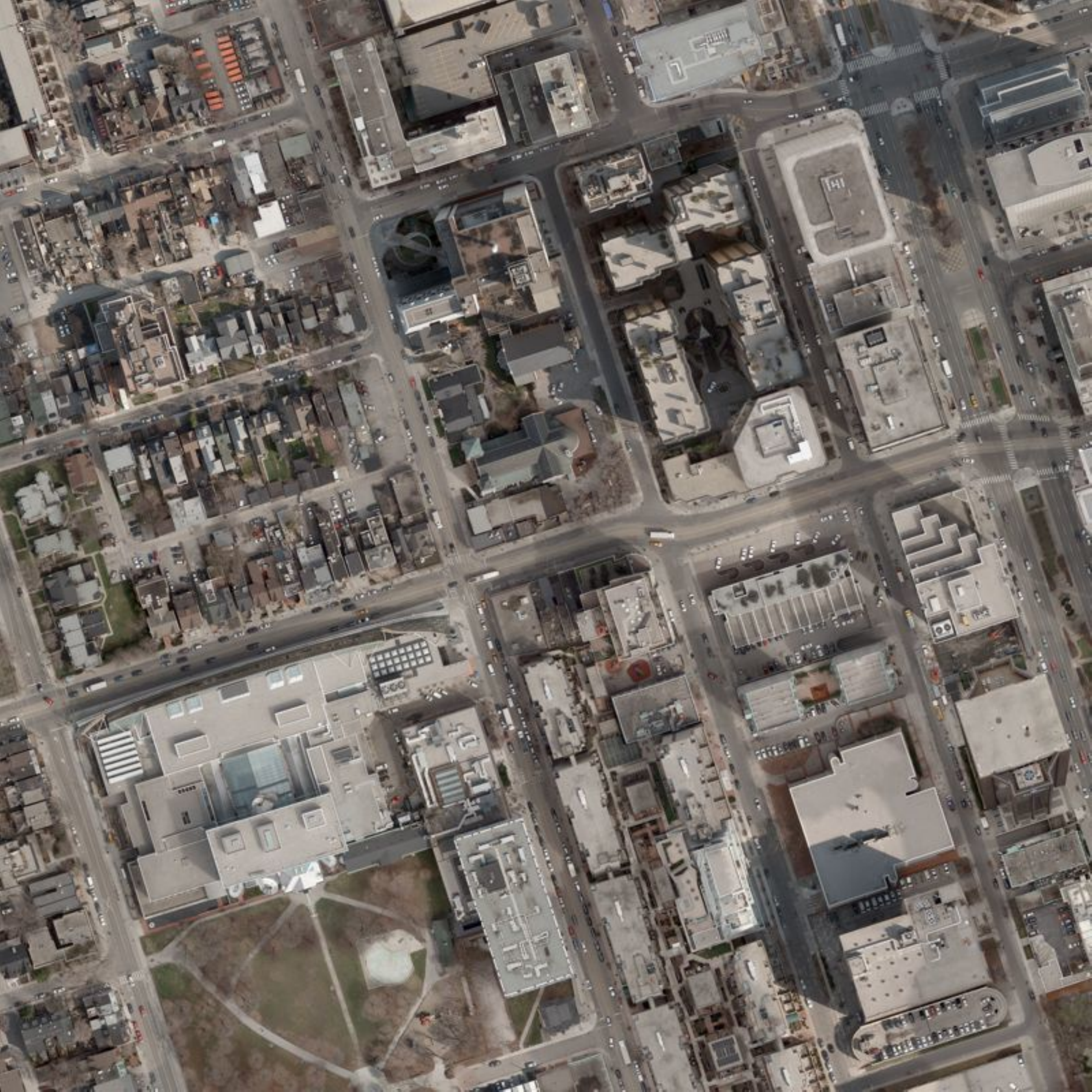}
  
  \vspace{0.5mm}
  
  \adjincludegraphics[width=\textwidth,trim={{.25\width} {.2\width} 0 {.2\width}},clip]{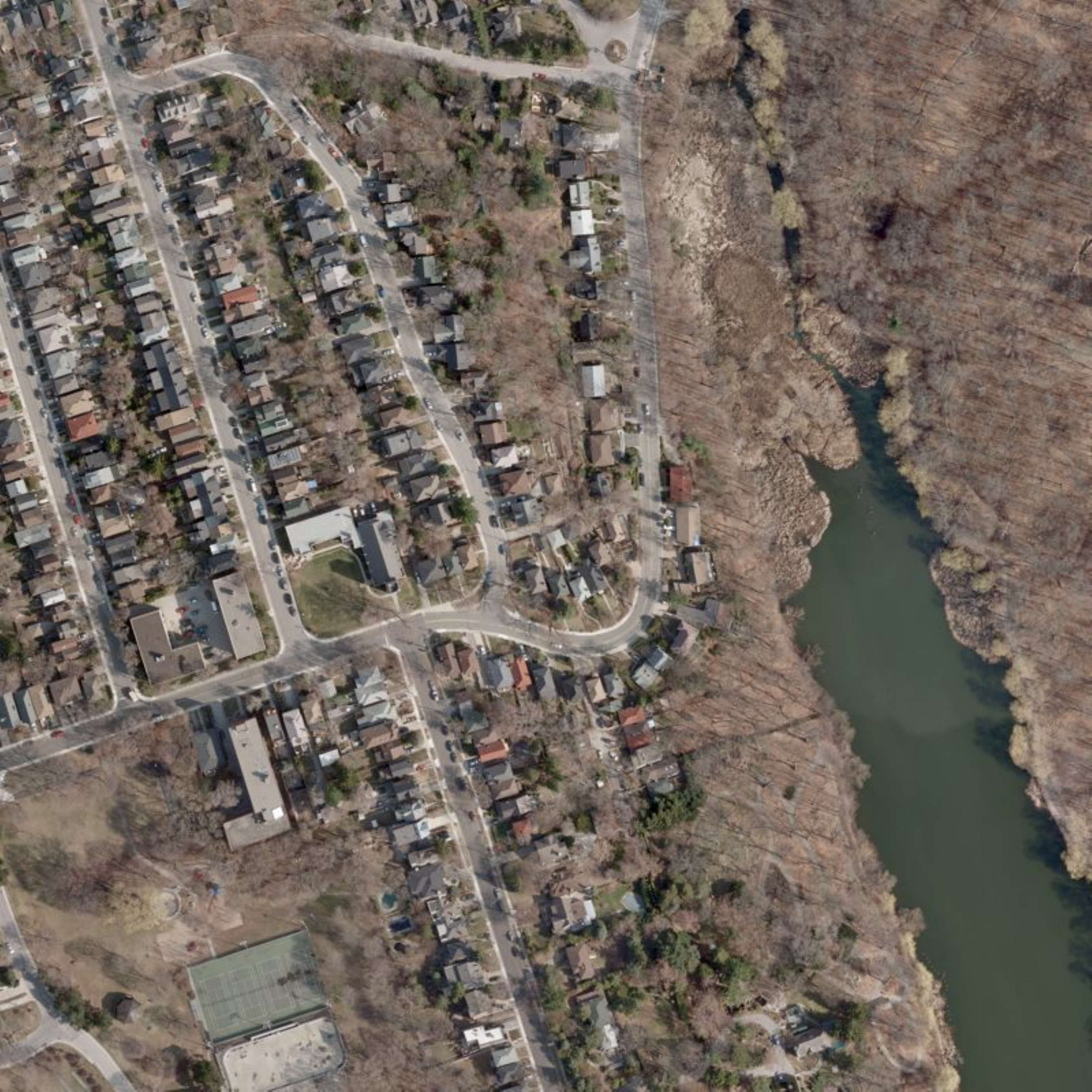}

 \caption{Input}
  \end{subfigure}
\begin{subfigure}[t]{0.16\textwidth}
  \adjincludegraphics[width=\textwidth,trim={ {.5\width} {.6\width} 0 0 },clip]{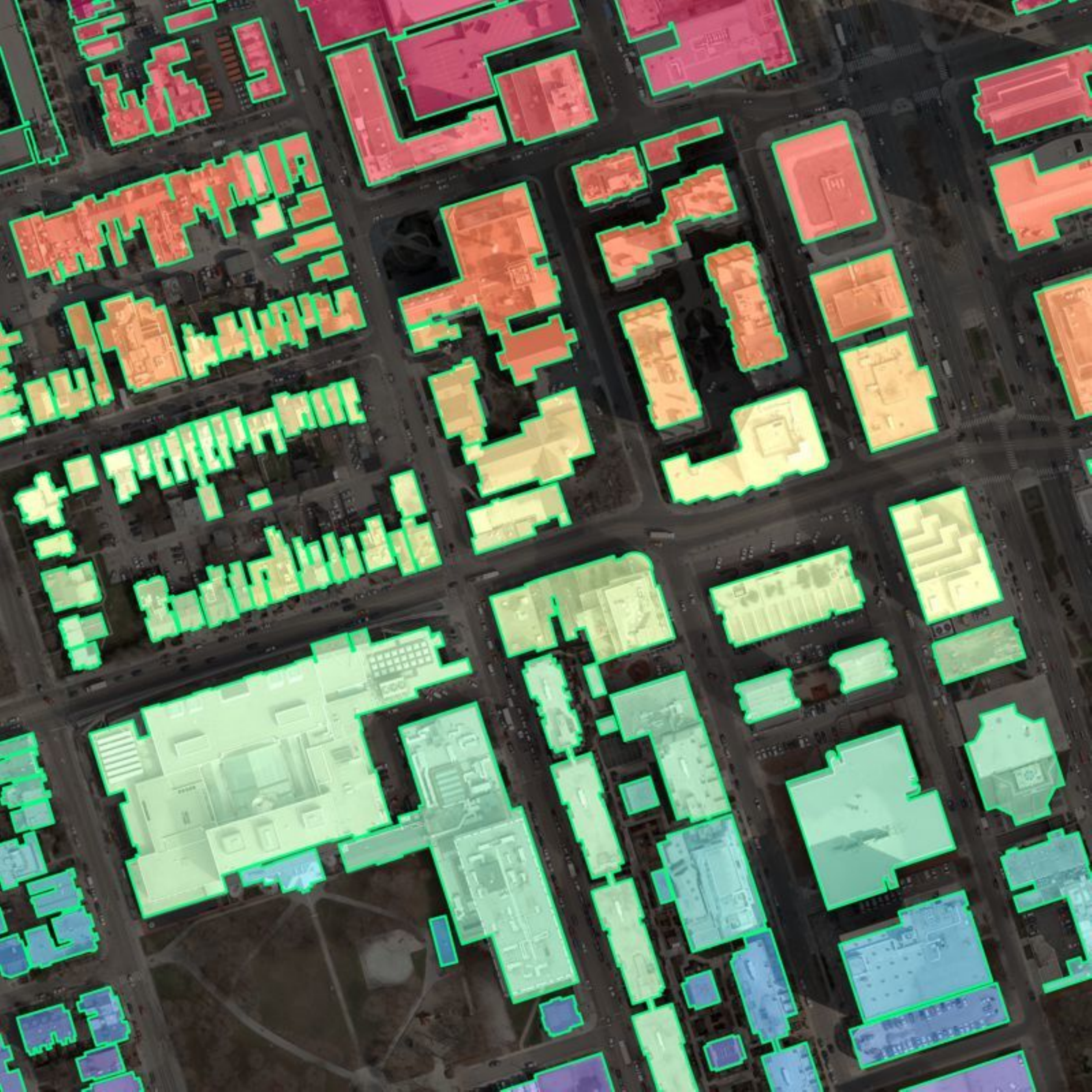}  
 
  \vspace{0.5mm}
   \adjincludegraphics[width=\textwidth,trim={{.25\width} {.2\width} 0 {.2\width}},clip]{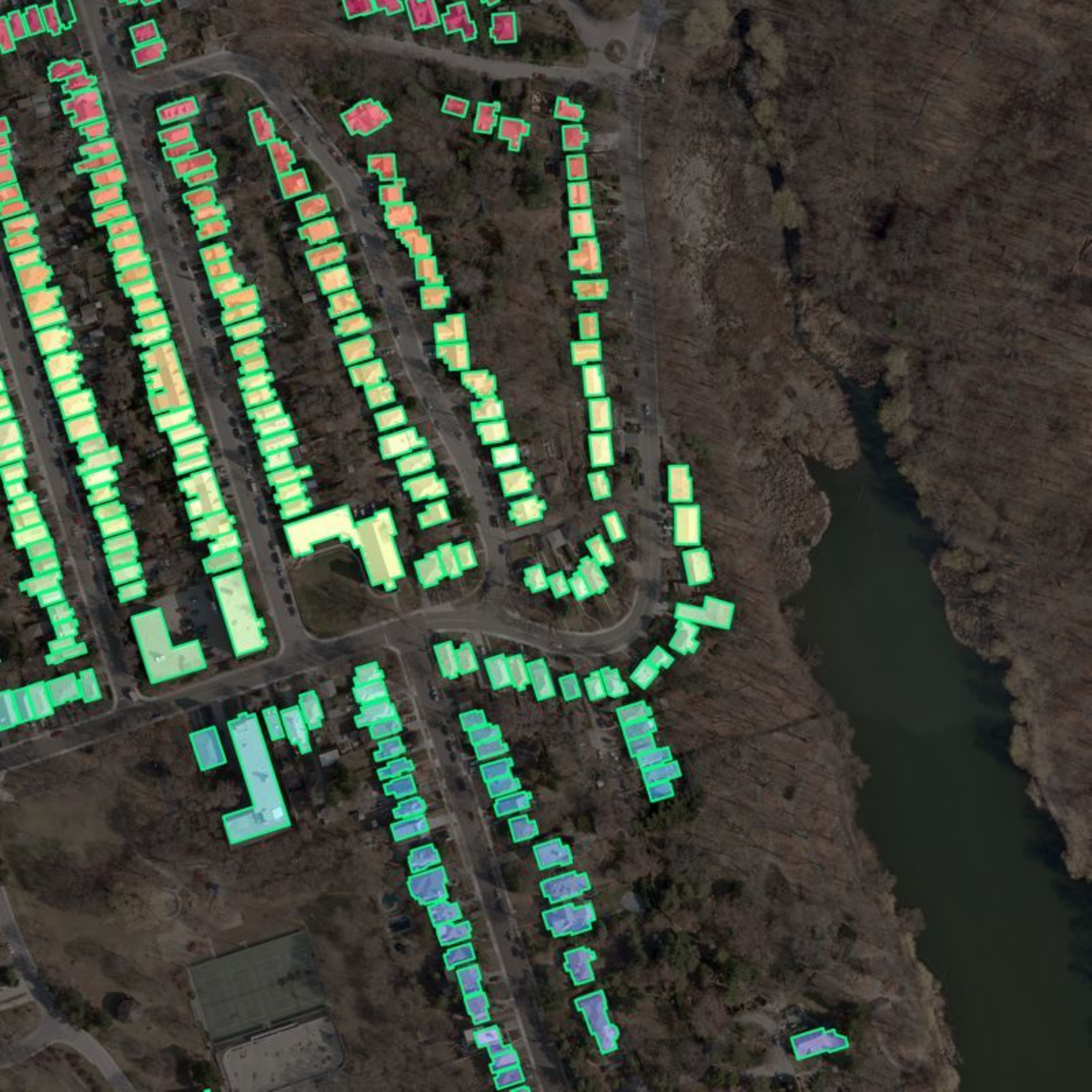}  
 \caption{GT}
 \end{subfigure}
 \begin{subfigure}[t]{0.16\textwidth}
  \adjincludegraphics[width=\textwidth,trim={ {.5\width} {.6\width} 0 0 },clip]{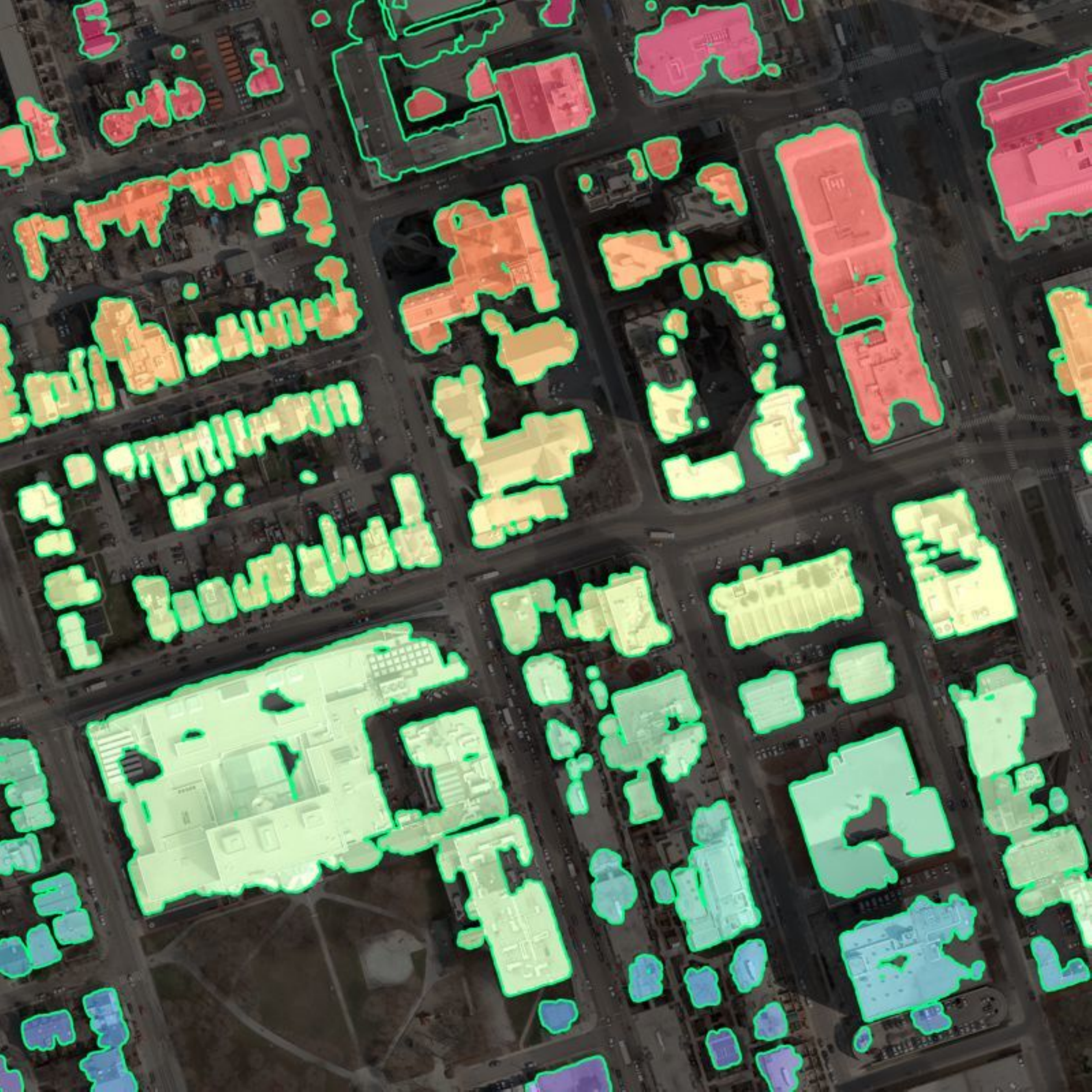} 
  
  \vspace{0.5mm}
  
 \adjincludegraphics[width=\textwidth,trim={{.25\width} {.2\width} 0 {.2\width}},clip]{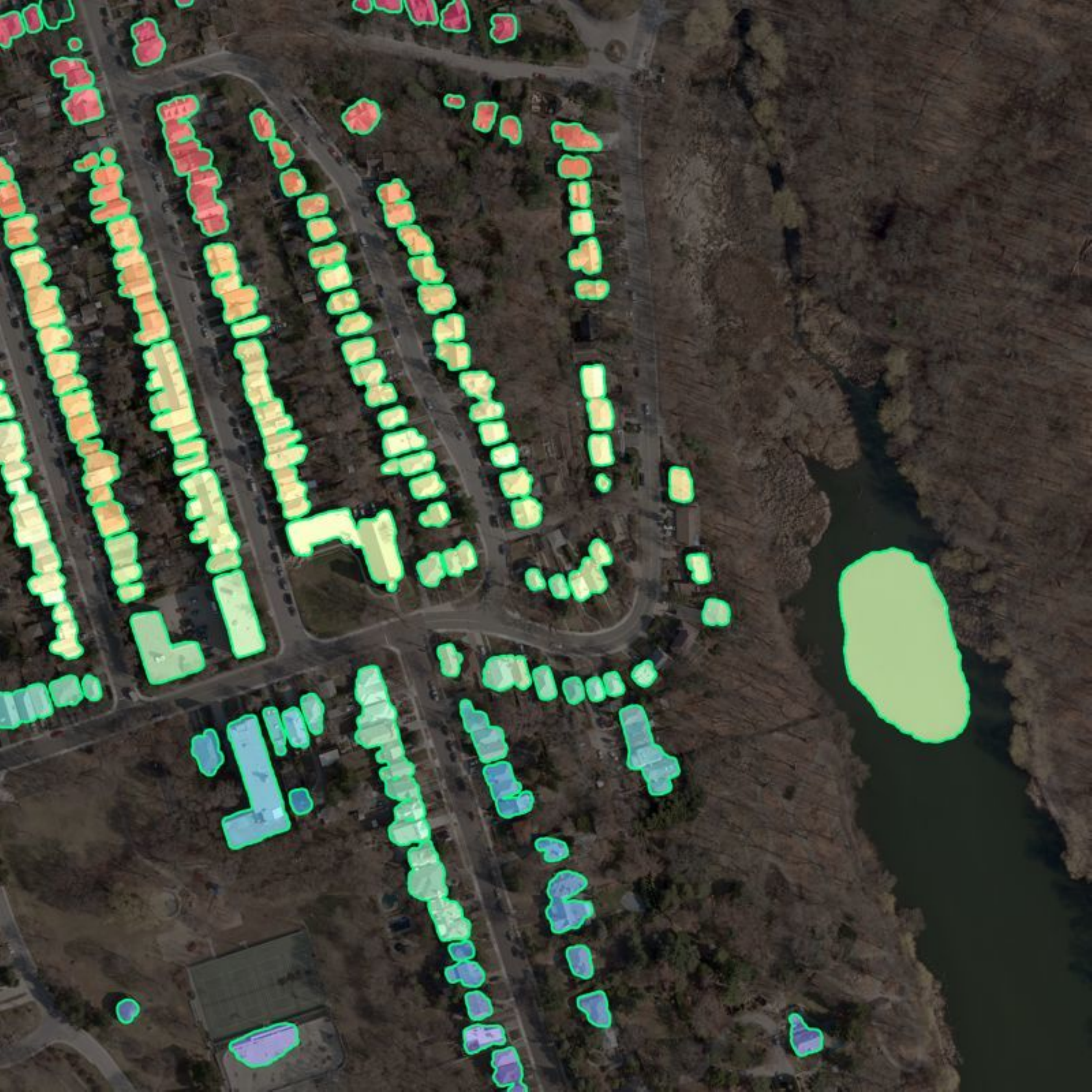}  
 \caption{ResNet56}
 \end{subfigure}
\begin{subfigure}[t]{0.16\textwidth}
   \adjincludegraphics[width=\textwidth,trim={ 0 {.6\width} {.5\width} 0},clip]{fig/instance_viz/623000_4833500.pdf}
  
  \vspace{0.5mm}
  
  \adjincludegraphics[width=\textwidth,trim={0   0  {.25\width}  {.4\width}},clip]{fig/instance_viz/629500_4834500.pdf}
 \caption{Input}
  \end{subfigure}
\begin{subfigure}[t]{0.16\textwidth}
   \adjincludegraphics[width=\textwidth,trim={0 {.6\width} {.5\width} 0},clip]{fig/instance_viz/623000_4833500_gt_instance.pdf}  
 
  \vspace{0.5mm}
  
  \adjincludegraphics[width=\textwidth,trim={0   0  {.25\width}  {.4\width}},clip]{fig/instance_viz/629500_4834500_gt_instance.pdf} 
 \caption{GT}
 \end{subfigure}
 \begin{subfigure}[t]{0.16\textwidth}
 \adjincludegraphics[width=\textwidth,trim={0 {.6\width} {.5\width} 0},clip]{fig/instance_viz/623000_4833500_resnet_opening_val_instance.pdf}  
  
  \vspace{0.5mm}
  
 \adjincludegraphics[width=\textwidth,trim={0   0  {.25\width}  {.4\width} },clip]{fig/instance_viz/629500_4834500_resnet_opening_val_instance.pdf} 
 \caption{ResNet56}
 \end{subfigure}
%
%
%
%
    \vspace{-3mm}
 \caption{Examples of building instance segmentation.}
 \label{fig:instance}
 \end{figure*}


\paragraph{Panoramas:} We  downloaded Google Streetview panoramas \cite{gsv} that densely populate the GTA. On average, we crawled around 520 full $360^{\circ}$ spherical panoramas for each $km^2$. In addition, we crawled the associated metadata, including the geolocation, address and the parameters of the spherical projection, including pitch, yaw and tilt angles.
\shenlong{We} 
resized all panoramas to 
$3200\times1600$ pixels. 

\paragraph{Aerial Imagery:} We use aerial images 
with full coverage of the GTA taken in 2009, 2011, 2012 and 2013.  
 They are orthorectified to $10cm$/pixel resolution for 2009 and 2011, and $5$ and $8cm$/pixel for 2012 and 2013 respectively. This contrasts satellite images, which are at best  $50cm$/pixel. 
 Our aerial images  have four channels (i.e., RGB and  Near infrared), and are 16 bit resolution for 2011 and 8 bit  for the rest. 
As is common practice in remote sensing \cite{utm}, we projected each image to the Universal Transverse Mercator (UTM) 17 zone in the WGS84 geodetic datum and tiled the area to $500\times500m^2$ images without overlap. 
Note that the images 
are not true orthophotos and thus facades are visible. 

\paragraph{Airborne LIDAR:} We also exploit  airborne LIDAR data captured in 2008 with a Leica ALS  sensor with a resolution of 6.8 points per $m^2$.
The total coverage  is 22 $km^2$. All of the points are also geo-referenced and projected to the UTM17 Zone in WGS84 geodetic datum.

\paragraph{Car setting:} Our recording platform includes a GoPro Hero 4 RGB camera, a Velodyne HDL-64E LIDAR  and a PointGray Bumblebee3 Stereo Camera mounted on top of the vehicle. All the sensors are calibrated and synchronized with a Applanix POS LV positioning system to record  real-time geo-location and orientation information. 
We have driven 
this platform for around 90km, which includes 
repeats of the same area. 
Note that we are  collecting and aligning new data from ground-view vehicles, and plan to have a much larger coverage 
by  the time of release.
 

\subsection{Maps as Annotations}

Manually labeling such a large scale dataset as TorontoCity is simply not possible. 
Instead, in this paper we exploit different sources of  high-precision maps covering the whole GTA to create our ground truth. 
Compared to online map services such as  OpenStreetMap \cite{osm} and Google Maps, our maps are much more accurate. 
Furthermore, they  contain many additional sources of detailed meta data which we exploit. 
One of the main challenges in creating TorontoCity was the alignment of the maps to all data sources.
In the following, we first  describe the annotated data composing TorontoCity and postpone our discussion on  the algorithms we  developed to align all data sources  to the next section.

\paragraph{Buildings:}
The TorontoCity dataset contains $400,000$ 3D buildings covering the full GTA. 
As shown in Fig. \ref{fig:dataset}, the buildings are very diverse, with the tallest being the CN Tower with 443m of elevation. 
Toronto contains many individual family houses, which makes tasks such as instance level segmentation particularly difficult. The mean height of each building is 4.7m, and the mean building area is $148m^2$. 
In contrast, the largest building has an area  $120,000m^2$. 
The level of detail of the 3D models varies per building (see \figref{fig:intro}). 
Many of these buildings are accurate to within centimeters 
and contain many other semantic primitives such as roof types, windows and balconies. 

\paragraph{Roads:}
Our maps contain  very accurate polylines representing streets, sidewalks, rivers and railways within the GTA. Each line segment is described with a series of attributes such as  name, road category and address number range. Road  intersections are explicitly encoded as intersecting points between polylines. 
Road curbs are also available, and describe the shape of  roads (see \figref{fig:intro}). 


\paragraph{Urban Zoning:}
Our maps contain 
government zoning information on the division of land into categories. 
This zoning 
includes   categories such as  residential, commercial, industrial and institutional. Note that multiple categories 
are allowed for one zone, e.g., commercial+residential. 
Understanding urban zoning is important in applications such as urban planning, real estate and law-enforcement. 


\paragraph{Additional data:} We also have  cartographic information with full coverage of the GTA. For instance, we have the location of all the poles, traffic lights, street lights and street trees with meta-information for each. 
The meta-information includes the height of the pole/traffic light, type of model of each street lights, trunk radius and species of each trees. We plan to exploit this in the near future. 

%% file: method.tex
 \begin{figure*}[t]
 \vspace{-0.7cm}
 \centering
 \includegraphics[width=0.32\linewidth]{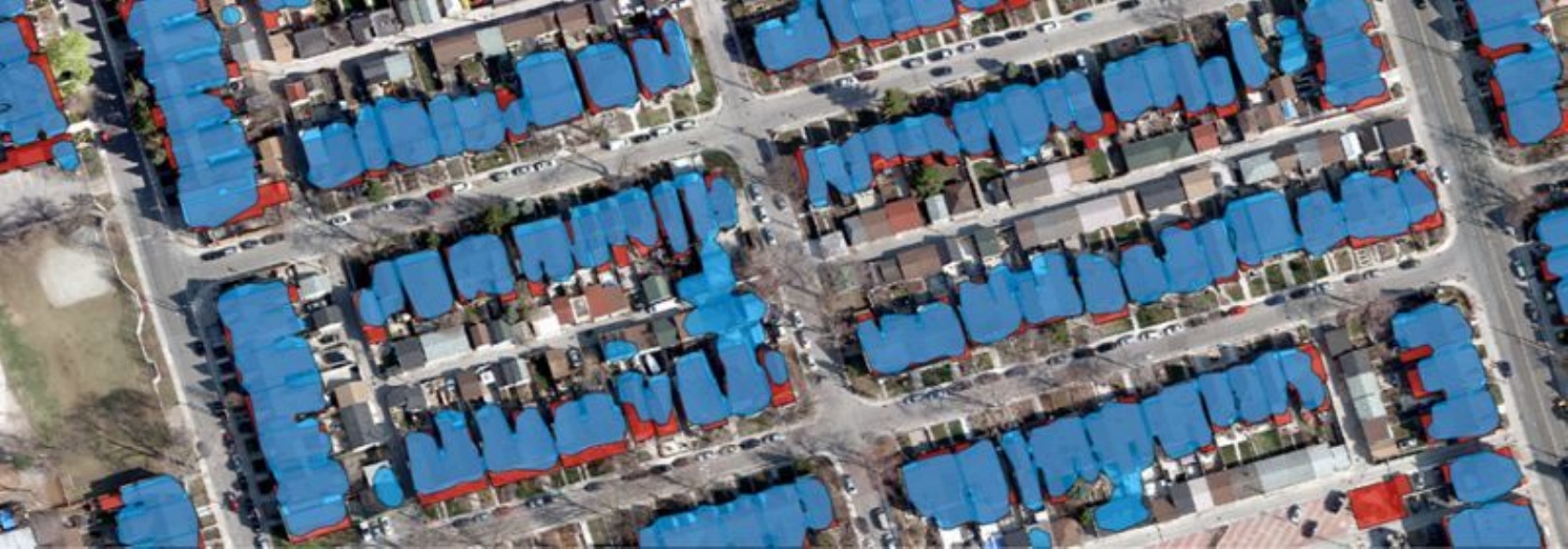}
  \includegraphics[width=0.32\linewidth]{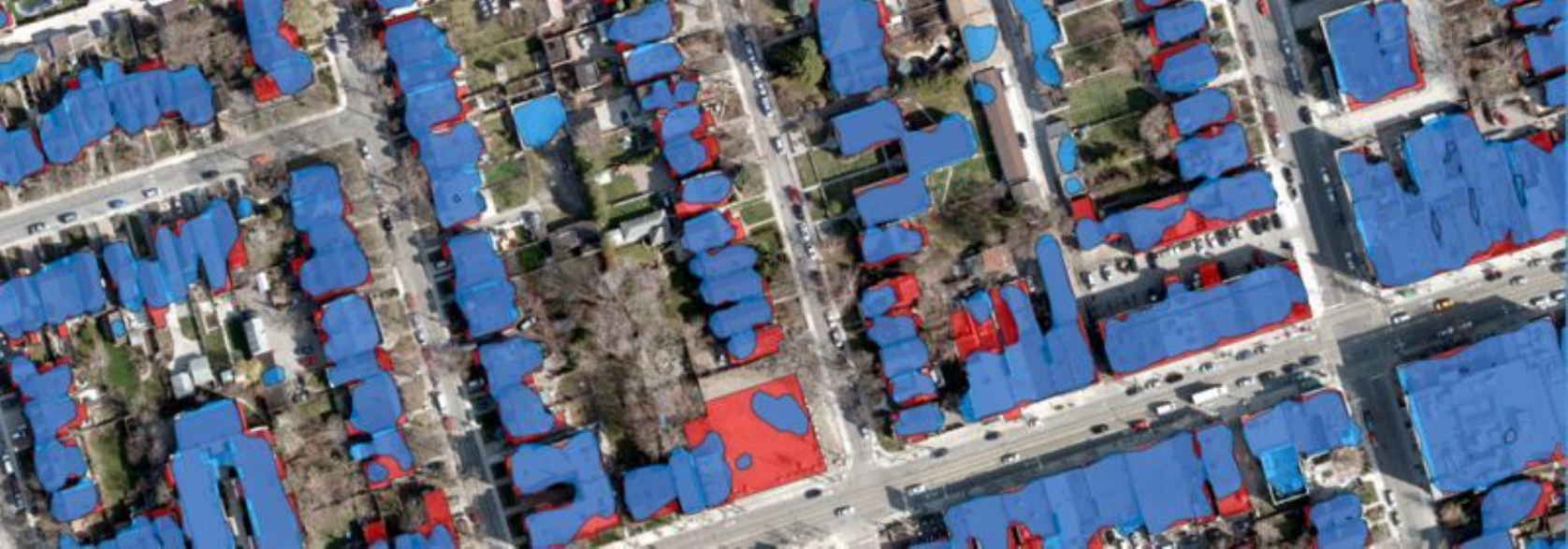}
   \includegraphics[width=0.32\linewidth]{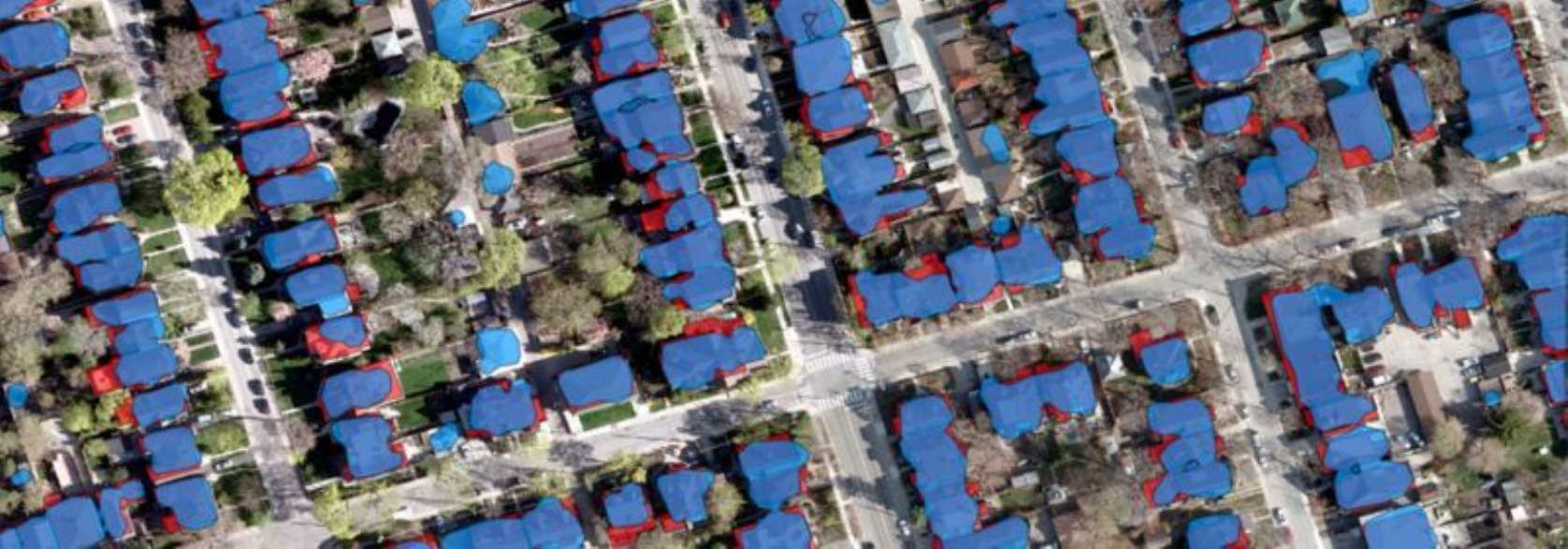}
       \vspace{-2mm}
 \caption{Qualitative results on building structured contour prediction: {\color{blue} ResNet} vs {\color{red} GT}}
 \label{fig:polygon}
     \vspace{-3mm}
 \end{figure*}

\section{Maps for Creating Large Scale Benchmarks}


In this section we describe our algorithms to automatically align 
maps with our  sources of imagery. We then describe the alignment of the different road maps. 

\subsection{Aligning Maps with All Data Sources}

The aerial images we employed are already perfectly aligned 
with the maps.   This, however, is not the case for the panoramas. 
As noted in \cite{housecraft}, the geolocalizaiton error is up to $5m$  with an average of $1.5m$, while rotation is very accurate. 
As a consequence, projecting our maps will not generate good ground truth due to the large misalignments as shown in Fig.~\ref{fig:ncc}. 
To handle this issue, we design an alignment algorithm that exploits both  aerial images and  maps. Their information is complementary, as aerial images give us appearance, while maps give us sparse structures (e.g., road curves). 

For this, we first rectify the panoramas by projecting them onto the ground-plane. We extract  a $400\times400$ m  ground plane region  with $10$cm/pixel resolution and  parameterize the alignment   with three degrees of freedom  representing  the camera's offset. 
We then perform a two step alignment process.  
We obtain a coarse 
alignment by maximizing a scoring function that compromises between appearance matching and a  regularizer. 
In particular, we use  normalized cross correlation (NCC) as our appearance matching term and  a Gaussian prior with mean $(0,0,2.5)m$ and diagonal covariance $(2, 2,0.2)m$. 
We re-scale  both aerial and ground images  to $[0, 1]$ before NCC.
 The solution space is a discrete search window in the range $[-10m, 10m]\times[-10m, 10m]\times[2.2m, 2.6m]$ with a step of $0.1m$.
We use exhaustive search to perform this search, and exploit the fact that NCC can be computed efficiently using FFT and  the Gaussian prior score is a fixed look-up-table. 
As shown in  Fig.~\ref{fig:ncc} this procedure produces very good coarse 
alignments. The alignment is coarse 
 as we reason at the aerial images' resolution, which is relatively lower. 

Our fine alignment then utilizes the road curves and aligns them to the boundary edges \cite{structedge} in the panorama. 
 We use a  search area of $[-1m, 1m]\times [-1m, 1m]$ with a step of 5cm. 
 This is followed by a human verification process that selects the images where this alignment succeeds. Mistakes in the alignment are due to occlusions (e.g., cars in the panoramas) as well as significant non-flat terrain.
 Our success rate is {34.35\%}, and it takes less than 2s to verify an image. In contrast annotating the alignment takes {20s}. 

 \begin{figure*}[t]
 \vspace{-0.7cm}
 \centering
\begin{subfigure}[t]{0.32\textwidth}
  \includegraphics[width=\textwidth]{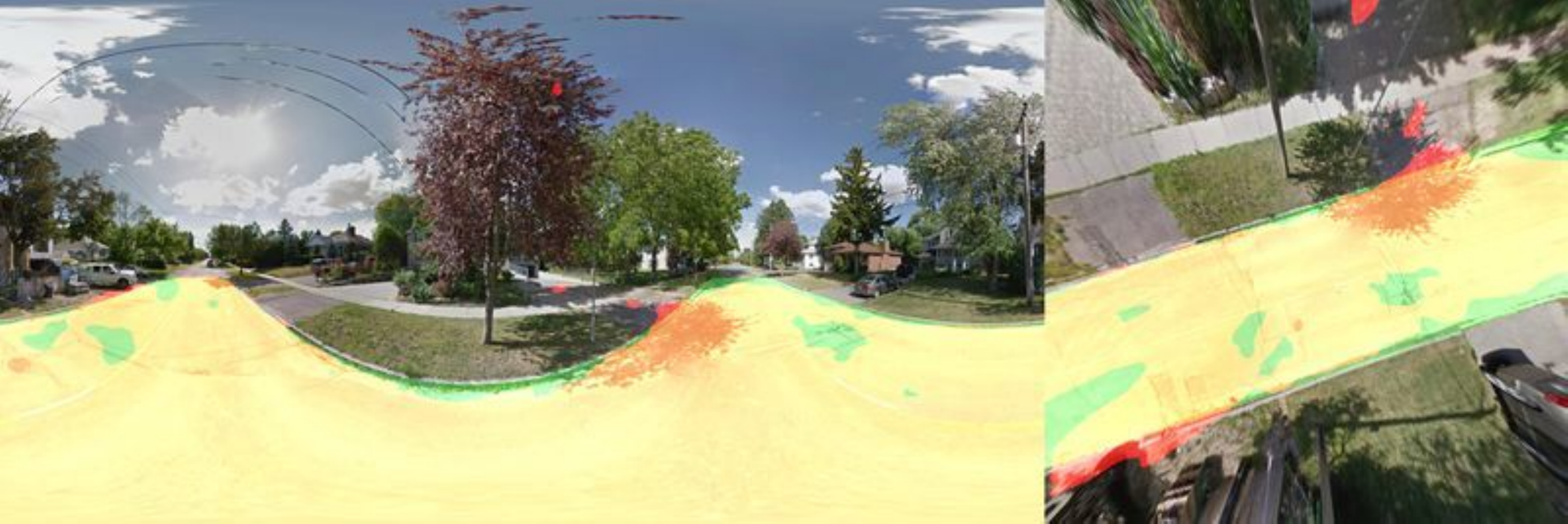}
  \end{subfigure}
\begin{subfigure}[t]{0.32\textwidth}
  \includegraphics[width=\textwidth]{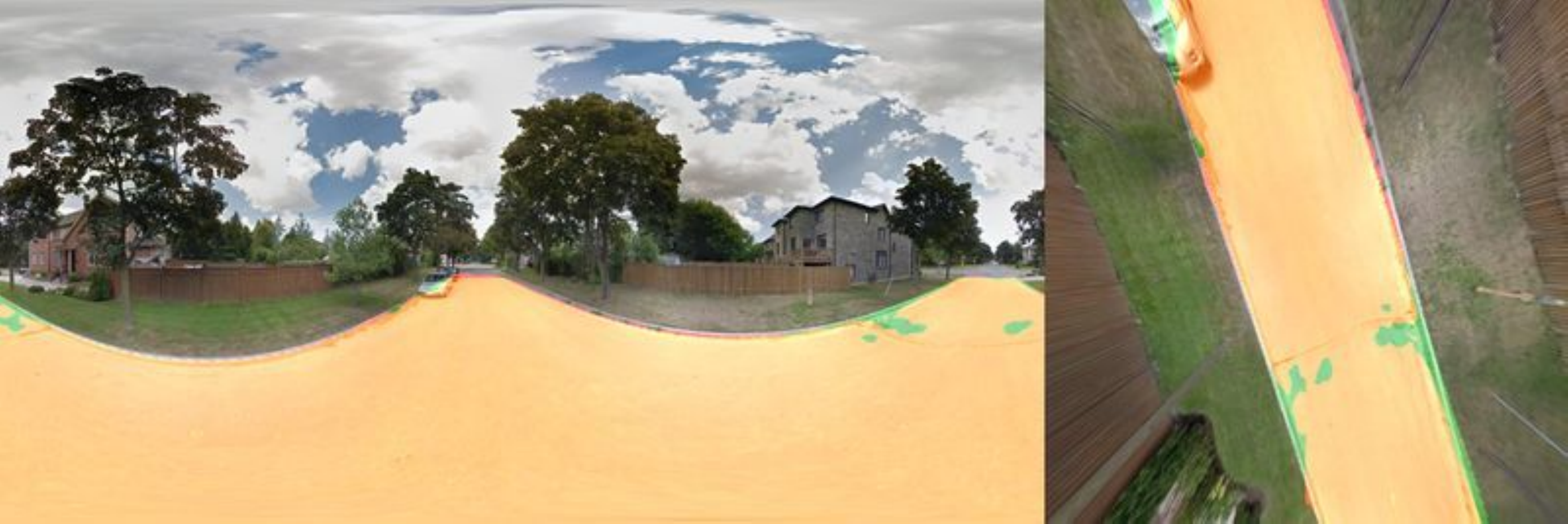} 
 \end{subfigure}
 \begin{subfigure}[t]{0.32\textwidth}
  \includegraphics[width=\textwidth]{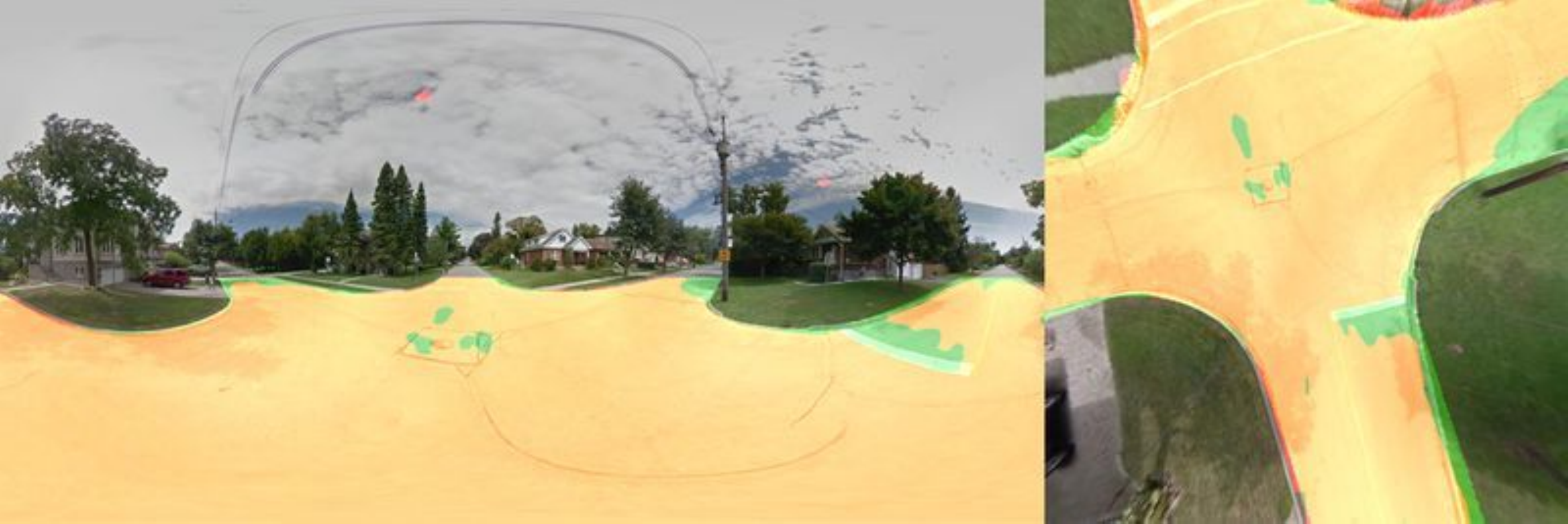} 
 \end{subfigure}
     \vspace{-2mm}
 \caption{Examples of road segmentation. Left: panoramic view; right: top-down view. (TP: {\color{yellow} yellow}, FP: {\color{red} red}, FN: {\color{green} green})}
 \label{fig:ground-seg}
     \vspace{-1mm}
 \end{figure*}
  
\begin{table}
\vspace{-0.3cm}
\begin{center}
\footnotesize	
\begin{tabular}{c|ccc}
Method & Mean & Building & Road\\ \hline
FCN \cite{fcn} & 77.64\% & \textbf{70.44\%} & 73.32\% \\ 
ResNet \cite{resnet} & \textbf{78.46\%} & 69.15\% & \textbf{76.44\%} \\ 
\hline
\end{tabular}
\end{center}
 \vspace{-6mm}
\caption{Aerial image semantic segmentation IoU. 
}
\label{tab:seg}
 \vspace{-2mm}
\end{table}%

\begin{table}
\vspace{-0.2cm}
\begin{center}
\footnotesize	
\begin{tabular}{c|cccc}
Method & WeightedCov & AP & Re-50\% & Pr-50\% \\ \hline
FCN & 39.74\%& 	8.04\% &	19.64\% &	 18.38\% \\ 
FCN + Open  & \textbf{43.19\%}&	16.45\% & 24.55\% & 36.09\% \\ 
ResNet  & 38.70\%&	10.47\% &	21.30\% &	 21.93\% \\ 
ResNet + Open  & 41.10\%& \textbf{22.92\%} & \textbf{22.78\%} & \textbf{43.78\%} \\ 
\hline
\end{tabular}
\end{center}
 \vspace{-6mm}
\caption{Building instance segmentation IoU. }
\label{tab:instance}
 \vspace{-3mm}
\end{table}%

 \subsection{Semantic Segmentation from Polyline Data}
 

Our maps provide two types of road  structures: curbs defining the road boundaries  as well as  center lines defining the connectivity (adjacency) in the street network. 
Unfortunately, these two sources are not aligned, and occasionally center lines are outside the road area. 
In this section we show our procedure of exploiting a Markov random field (MRF) to align road centerlines and curves. We can then generate the polygons describing the road surfaces. 
Fig.  \ref{fig:roadSurfaceGeneration} shows an example for the road surface generation.

Let  $y_i \in \{0,1, \cdots, k\}$ be the assignment of the $i$-th curb segment to one of the $k$ nearest centerline segments, where state $0$  denotes no match. 
We define an MRF composed of unary and pairwise terms, which connects 
only adjacent curbs segments, and thus naturally form a set of chains. 
For the 
unary terms $\phi_{un}(y_i)$, 
we use the weighted sum of the  distance of the curve to each  centerline segment (condition on the state) and the angular distance between curves and centerlines. 
For the 
pairwise terms $\phi_{con}(y_i,y_{i+1})$, we employ a Potts potential that encourages smoothness along the road. 
This is important as otherwise there may be holes in places such as intersections, 
since the center of the intersection is further away from 
other points.  
Due to the chain structure of the graphical model, inference can be done exactly and efficiently in parallel for each chain using dynamic programming. 
Our formulation allows for multiple curbs to be matched to one road, which is needed as there are curbs on both sides of the centerline. 
We manually inspect the results and mark errors as ``don't care'' regions. 
We convert each continuous curb-road center line assignment to polygons which gives us the final road surface. We refer the reader to Fig. \ref{fig:roadSurfaceGeneration} for an example.

%% file: benchmarks.tex
\begin{table*}
\begin{center}
\footnotesize	
\begin{tabular}{c|cccccc|cccccc}
 & \multicolumn{6}{c}{Road centerline} &  \multicolumn{6}{|c}{Road curb} \\ 
Method & F1$_{\mathrm{0.5}}$ & Pr$_{\mathrm{0.5}}$ & Re$_{\mathrm{0.5}}$ & F1$_{\mathrm{2}}$ & Pr$_{\mathrm{2}}$ & Re$_{\mathrm{2}}$ & F1$_{\mathrm{0.5}}$ & Pr$_{\mathrm{0.5}}$ & Re$_{\mathrm{0.5}}$ & F1$_{\mathrm{2}}$ & Pr$_{\mathrm{2}}$ & Re$_{\mathrm{2}}$ \\ \hline
FCN 			& 0.169 & 0.156 & 0.186 & 0.626 & 0.576 & \textbf{0.687} & 0.444 & 0.413 & 0.482 & 0.778 & 0.726 & \textbf{0.837} \\
FCN+Close 		& \textbf{0.173} & 0.164 & 0.183 & 0.639 & 0.604 & 0.678 & 0.444 & 0.427 & 0.462 & 0.781 & 0.752 & 0.812 \\
ResNet 			& 0.162 & 0.143 & \textbf{0.186} & 0.613 & 0.567 & 0.667 & \textbf{0.575} & 0.585 & \textbf{0.566} & 0.796 & 0.830 & 0.765 \\
ResNet+Close 	& 0.162 &\textbf{ 0.169} & 0.155 & \textbf{0.644} & \textbf{0.671} & 0.619 &0.568 & \textbf{0.614} & 0.529 & \textbf{0.799} & \textbf{0.862} & 0.745 \\ \hline
\end{tabular}
\end{center}
 \vspace{-6mm}
\caption{Road centerline and curb results. Metric: F1, Precision, Recall with minimal distance threshold 0.5m and 2m.}
\label{tab:centerline}
 \vspace{-2mm}
\end{table*}%
 \section{Benchmark Tasks and Metrics}
 

We designed a diverse set of benchmarks to push  computer vision approaches to reason about  geometry,  semantics  and grouping. To our knowledge, no previous dataset is able to do this at this scale. 
In the evaluation server, participants can submit results using any subset of the imagery types provided in the benchmark (e.g., aerial images, panoramas, Go-Pro, stereo). 
In this section, 
we briefly describe the tasks and metrics, and 
refer the reader to the supplementary material for further 
details.  Note also that Fig. \ref{fig:intro} shows  an illustration of some of our tasks. 



\paragraph{Building Footprint and Road segmentation:} Our first task is semantic segmentation of building footprints and roads. 
Following common practice in semantic segmentation, we utilize mean Intersection-Over-Union (mIOU) as our metric. This is evaluated from a top-down view. 


\paragraph{Building Footprint Instance Segmentation:} Our second task is  building instance segmentation. 
We adopt multiple metrics for this task, since there is no consensus in the community of what is the best metric.  We thus evaluate weighted coverage (Cov), average precision (AP) as well as instance level precision and recall at $50\%$. 

\paragraph{Building Structured Contours:} 
Most semantic and instance  segmentation algorithms produce "blob"-like results, which do not follow the geometry of the roads and/or buildings. 
We thus want to push the community to produce instance segmentations that follow the structure of the primitives. 
Towards this goal, we define a metric  that  merges (in a multiplicative fashion)   segmentation scoring  with geometric similarity. In particular, segmentation is measured in terms of IOU, and we exploit the similarity between the turning functions of the estimated and ground truth polygons as a geometric metric. 
We refer the reader to the supplementary material for more details. 

\paragraph{Road Topology:} One of the remaining fundamental challenges in mapping  is  estimating road topology. In this task, participants are asked to extract  polylines that represent road curbs and road centerlines  in bird's eye perspective. 
We discretize both estimated and ground truth polylines in intervals  of size $10$cm. 
We  define precision and recall as our metrics, where an 
estimated segment is correct if its distance to the closest segment on the target polyline set is smaller than a threshold (i.e., 0.5m and 2.0m). 

\begin{table*}
\vspace{-0.7cm}
\begin{center}
\footnotesize	
\begin{tabular}{c|cccc|cccc}
Method & AlexNet \cite{alexnet} & VGG-16 \cite{vgg} & GoogleNet \cite{googlenet}  & ResNet-152 \cite{resnet}  & AlexNet \cite{alexnet} & ResNet-32 \cite{resnet}  & GoogleNet \cite{googlenet}  & NiN \cite{nin}  \\  \hline 
From-scratch & no & no  & no & no   & yes & yes   & yes & yes    \\
Top-1 accuracy & 75.49\% & 79.12\% & 77.95\% & \textbf{79.33\%} & 66.48\%   & 75.65\%   & 75.08\%  & \textbf{79.07\%} \\
\hline
\end{tabular}
\end{center}
 \vspace{-6mm}
\caption{ Ground-Level Urban Zoning Classification}
\label{tab:zone_seg}
 \vspace{-2mm}
\end{table*}%

\paragraph{Ground Road Segmentation:}
We use IOU as our metric. 

\paragraph{Ground Urban Zoning Classification:}
This benchmark is motivated by the human's ability to recognize the urban function of a local region  by its appearance.  
We use Top-1 accuracy as our metric and evaluate on the ground view. 

\paragraph{Urban Zoning Segmentation:} Our goal is to produce a segmentation in bird's eye view of the  different urban zones including residential, commercial, open space, employment, \etc. 
We utilize IOU as our metric. 

\paragraph{Building Height Estimation:}

 This tasks consists on estimating building height. Useful cues include size of the buildings, pattern of shading and shadows as well as the imperfect rectification in aerial views. 
We adopt root mean square error in the log domain (log-RMSE) as our metric.  

\paragraph{Additional Tasks:} We plan to add many  tasks in the coming  months. This includes detecting trees and recognizing their species. Moreover, the accurate 3D building models allow us to build a benchmark of normal estimation as well as facade parsing.  We also plan to have benchmarks for detection and segmentation of traffic lights, traffic signs and poles. We 
are just scratching the surface of the plenthora of possibilities with this dataset. 

%
 
%

%% file: results.tex
 \section{Experimental Evaluation} 
 
\begin{table}[t]
\vspace{-0.3cm}
\begin{center}
\footnotesize	
\begin{tabular}{c|cc}
Method & WeightedCov &  PolySim \\ \hline
FCN 	& \textbf{0.456} & \textbf{0.323} \\
ResNet 	& 0.401 & 0.292 \\
\hline
\end{tabular}
\end{center}
 \vspace{-6mm}
\caption{Building contour results. }
\label{tab:contour}
 \vspace{-2mm}
\end{table}%

\begin{table}[t]
\begin{center}
\footnotesize	
\begin{tabular}{c|ccc}
Method & Residential & Open Space & Others \\ \hline
FCN 	& \textbf{60.20\%} & 32.20\% & \textbf{5.57\%} \\
ResNet 	& 51.71\% & \textbf{33.63\%} & 1.49\% \\
\hline
\end{tabular}
\end{center}
 \vspace{-6mm}
\caption{Qualitative results for urban zoning segmentation.}
\label{tab:zone_seg}
 \vspace{-2mm}
\end{table}%
 
We perform a pilot study of the difficulty of our tasks in a  subset of TorontoCity, containing 125 $km^2$ region (50 $km^2$ for training, 50 $km^2$ for testing and $25 km^2$ for validation). The train/val/test regions do not  overlap 
and are not adjacent. We utilize 56K streetview images around these regions (22K for training, 18K for validation and 16K for testing
). Hyper-parameters are chosen based on validation performance, and all numbers reported are on the testing set. 

To perform the different segmentation related tasks, we train two types of convolutional networks: a variant of  FCN-8 architecture \cite{fcn} as well as a ResNet \cite{resnet} with 56 convolutional layers. More details are in supp. material. 


\paragraph{Semantic Segmentation:}
As shown in  Tab.~\ref{tab:seg}, both networks perform well. 
Fig.~\ref{fig:aerial-seg} illustrates  qualitative results of ResNet56 output. It is worth noting that large networks such as  ResNet56 can be trained from scratch given our large-scale dataset. Visually ResNet's output tends to be more sharp, while FCN's output is more smooth. 
 
\paragraph{Instance Segmentation:} We estimate instance segmentation  by taking the output of the semantic labeling and performing connected-component labeling. 
Each component is assigned a different label. 
Since convolutional nets   tend to generate blob like structures, a single component might contain  multiple instances connected with a small number of pixels. To alleviate this problem, we apply   morphological opening operators over the semantic labeling masks (an erosion filtering followed by a dilation filtering with the same size). As shown in Tab.~\ref{tab:instance} and Fig.~\ref{fig:instance} the performance is low. 
There is still much for the community to do to solve this task.  
With more than $400,000$ buildings, the TorontoCity dataset provides an ideal 
platform for new developments.  


\paragraph{Road Centerlines and Curbs:}
We compute the medial axis of the semantic segmentation  to extract the skeleton of the mask as our estimate of road centerline. In order to smooth the skeletonization, we first conduct a morphological closing operator (dilation followed by erosion) over the road masks. 
To estimate road curbs, 
we simply extract the contours of the  road segmentation and exploit closing operator. 
As shown in Table.~\ref{tab:seg}, ResNet achieves the highest score in both tasks, and morphological filtering helps for both networks. 
Qualitative results are shown in Fig.~\ref{fig:aerial-seg}. 
Note that there is still much room for improvement. 

\paragraph{Building Contours:} We compute  building contours from our estimated  building instances, and apply the Ramer-Douglas-Peucker algorithm \cite{rdp} to simplify each polygon with a threshold of $0.5$m. This results in polygons with  13 vertices on average. 
As shown in  Tab.~\ref{tab:contour} and Fig. \ref{fig:polygon}, this simple procedure offers reasonable yet not satisfactory resulst. This suggests there is still a large improvement space for generating building polygons from aerial images. 

\paragraph{Ground Urban Zoning Classification:}
We train multiple state-of-the-art convolutional networks for this task including  AlexNet \cite{alexnet}, VGG-16 \cite{vgg}, GoogleNet \cite{googlenet} and ResNet-152 \cite{resnet} that are fine-tuned from the ImageNet benchmark \cite{imagenet}. We also train AlexNet \cite{alexnet}, ResNet-32 \cite{resnet}, Network-In-Network \cite{nin} and ResNet-152 \cite{resnet} from scratch over our ground-view panoramic image tiles. As shown in Table.~\ref{tab:seg} ResNet-152 with pre-trained initialization achieves the best results. Net-in-net achieves the best performance among all  models that are trained from scratch. For more details, please refer to the supplementary material. 

\paragraph{Urban Zoning Segmentation:}
This is an extremely hard task from aerial views alone. 
To simplify it,  we merged the zone-types into residential, others (including commercial, utility and employment) as well as open spaces (including natural, park, recreational \etc ). Please refer to the supplementary material for detailed label merging. As shown in  Tab.~\ref{tab:zone_seg} more research is needed to solve this task. 

\paragraph{Ground-view road segmentation:}
We utilize a subset of the labeled panoramas, which includes 1000 training, 200 \joeljustin{validation} 
and 800 \joeljustin{testing images.} 
The average IOU is $97.21\%$. The average pixel accuracy is $98.64\%$ and average top-down IOU is $87.53\%$. This shows that a state-of-the-art neural network can nearly 
solve this task, suggesting that it is promising to automatically generate high-resolution maps by capturing geo-referenced street-view panoramas. 

\paragraph{Building Height:} No network was able to estimate building height from aerial images alone. This task is either too hard, or more sophisticated methods are needed. For example, utilizing ground imagery seems a logical first step.

%
%


%% file: conc.tex
 \section{Conclusions} 
 
In this paper, we have argued that the field is in need of large scale benchmarks that will allow joint reasoning about geometry, grouping and 
semantics. Towards this goal, we have created the TorontoCity benchmark,  covering the full Greater Toronto area (GTA) with $712.5km^2$ of land, $8439km$ of road and around $400,000$ buildings. Unlike existing datasets, our benchmark provides a wide variety of views of the world captured from airplanes, drones, as well as cars driving around the city. \joeljustin{As} 
using human annotators is not feasible for such a large-scale dataset, we have exploited different sources of  high-precision maps  to create our ground truth.
  We have designed a wide variety of tasks  including   building height estimation, road centerline and curb extraction, building instance segmentation,  building contour extraction (reorganization), semantic labeling and scene type classification (recognition). 
Our pilot study shows that most of these tasks are still difficult for modern convolutional networks. 
{We plan to extend the current set of benchmarks with tasks such as building reconstruction, facade parsing, tree detection, tree species categorization, traffic light detection, and traffic sign detection.}
This is only the beginning of the exciting TorontoCity benchmark.